\documentclass[]{opendatalab}
\usepackage{caption}
\usepackage{xcolor}
\usepackage{longtable}
\usepackage{listings}
\usepackage[numbers]{natbib}
\usepackage{amssymb}
\usepackage{tabularx}
\usepackage{float}
\usepackage{graphicx}
\usepackage{makecell}
\usepackage[table]{xcolor}
\usepackage{cleveref}
\usepackage{array}
\usepackage[misc]{ifsym}

\definecolor{codegreen}{rgb}{0,0.6,0}
\definecolor{codegray}{rgb}{0.5,0.5,0.5}
\definecolor{codepurple}{rgb}{0.58,0,0.82}
\definecolor{backcolour}{rgb}{0.95,0.95,0.92}
\definecolor{promptcolor}{HTML}{D1D0F2}
\definecolor{promptcolorheader}{HTML}{bdbcec}
\makeatletter
\newcommand\andauthor{%
  \g@addto@macro\authorlist{\par\vspace{1mm}}% Inserts a paragraph break and some vertical space
}
\newcommand\nlauthor[2][]{%
  \addtolist[#1]{#2}{\authorlist}{\authorformat}{}% The separator is now empty!
}
\makeatother

\RequirePackage{xspace}
\makeatletter
\DeclareRobustCommand\onedot{\futurelet\@let@token\@onedot}
\def\@onedot{\ifx\@let@token.\else.\null\fi\xspace}

\makeatother

\definecolor{codegreen}{rgb}{0,0.6,0}
\definecolor{codegray}{rgb}{0.5,0.5,0.5}
\definecolor{codepurple}{rgb}{0.58,0,0.82}
\definecolor{backcolour}{rgb}{0.95,0.95,0.92}
\definecolor{promptcolor}{HTML}{D1D0F2}
\definecolor{promptcolorheader}{HTML}{bdbcec}
\newcommand{\promptbox}[2]{
\begin{tcolorbox}[
top=0.3em,bottom=0.3em,left=0.5em,right=0.5em,
toptitle=0.3em,bottomtitle=0.2em,boxsep=0pt,
colframe=promptcolorheader,colback=promptcolor!50,boxrule=0.5pt,
]
\footnotesize
\end{tcolorbox}
}
\lstdefinestyle{mystyle}{
    backgroundcolor=\color{backcolour},   
    commentstyle=\color{codegreen},
    keywordstyle=\color{magenta},
    numberstyle=\tiny\color{codegray},
    stringstyle=\color{codepurple},
    basicstyle=\ttfamily\footnotesize,
    breakatwhitespace=false,         
    breaklines=true,                 
    captionpos=b,                    
    keepspaces=true,                 
    numbers=left,                    
    numbersep=5pt,                  
    showspaces=false,                
    showstringspaces=false,
    showtabs=false,                  
    tabsize=2
}

\title{RxnCaption: Reformulating Reaction Diagram Parsing as Visual Prompt Guided Captioning}
\vspace{10pt}

\author[1,2*]{Jiahe Song}
\author[3,1*]{Chuang Wang}
\author[4,1*]{Bowen Jiang}
\author[1*]{Yinfan Wang}
\author[1,5*]{Hao Zheng}
\andauthor
\nlauthor[1]{Xingjian Wei}
\author[6]{Chengjin Liu}
\author[3,1]{Rui Nie}
\author[1]{Junyuan Gao}
\author[1]{Jiaxing Sun}
\author[1]{Yubin Wang}
\andauthor
\nlauthor[1]{Lijun Wu}
\author[5\ \textrm{\Letter}]{Zhenhua Huang}
\author[1\ \textrm{\Letter}]{Jiang Wu}
\author[3\ \textrm{\Letter}]{Qian Yu}
\author[1\ \textrm{\Letter}]{Conghui He}

\affiliation[1]{Shanghai AI Laboratory}
\affiliation[2]{Shanghai Jiao Tong University}
\affiliation[3]{Beihang University}
\affiliation[4]{Peking University}
\affiliation[5]{South China Normal University}
\affiliation[6]{Northwestern Polytechnical University}
\abstract{
   Large-scale chemical reaction datasets are crucial for AI research in chemistry. However, existing chemical reaction data often exist as images within papers, making them not machine-readable and unusable for training machine learning models. In response to this challenge, we propose the \textbf{RxnCaption} framework for the task of chemical Reaction Diagram Parsing (RxnDP). Our framework reformulates the traditional coordinate prediction driven parsing process into an image captioning problem, which Large Vision-Language Models (LVLMs) handle naturally.
We introduce a strategy termed ``\emph{BBox and Index as Visual Prompt}'' (BIVP), which uses our state-of-the-art molecular detector, MolYOLO, to pre-draw molecular bounding boxes and indices directly onto the input image. This turns the downstream parsing into a natural-language description problem. Extensive experiments show that the BIVP strategy significantly improves structural extraction quality while simplifying model design.
We further construct the \texttt{RxnCaption-15k} dataset, an order of magnitude larger than prior real-world literature benchmarks, with a balanced test subset across four layout archetypes. Experiments demonstrate that RxnCaption-VL achieves state-of-the-art performance on multiple metrics.
We believe our method, dataset, and models will advance structured information extraction from chemical literature and catalyze broader AI applications in chemistry. We will release data, models, and code on GitHub.
}

\date{\today}
\correspondence{Zhenhua Huang, \email{huangzhenhua@m.scnu.edu.cn}, Jiang Wu, \email{wujiang@pjlab.org.cn}, Conghui He, \email{heconghui@pjlab.org.cn},  Qian Yu, \email{qianyu@buaa.edu.cn}} 

\metadata[Dataset]{\url{https://huggingface.co/datasets/songjhPKU/RxnCaption-15k}}
\begin{document}

\maketitle

\section{Introduction}

Machine learning excels in predicting chemical reaction outcomes~\cite{schwaller2019molecular}, optimizing conditions~\cite{de2019synthetic}, and planning retrosynthetic routes~\cite{zhong2024recent}. However, progress is limited by the scarcity of large, diverse datasets~\cite{ramos2025review,zhong2024recent,ding2025survey}. Public datasets like USPTO-50k~\cite{schneider2016s} lack the size and diversity of commercial databases such as SciFinder~\cite{gabrielson2018scifinder} and Reaxys~\cite{lawson2014making}. Additionally, much valuable data is trapped in unstructured image formats in scientific literature, hindering machine learning training.
Automated deep learning approaches offer a promising solution, with significant progress in molecular structure image recognition~\cite{qian2023molscribe} and text-based reaction extraction~\cite{zhong2023reaction}. The use of Large Visual Language Models (LVLMs)~\cite{wang2025gtr} has accelerated advancements in this area.

\begin{figure*}[t]
    \centering
    \vspace{-5pt}
    \includegraphics[width=\textwidth]{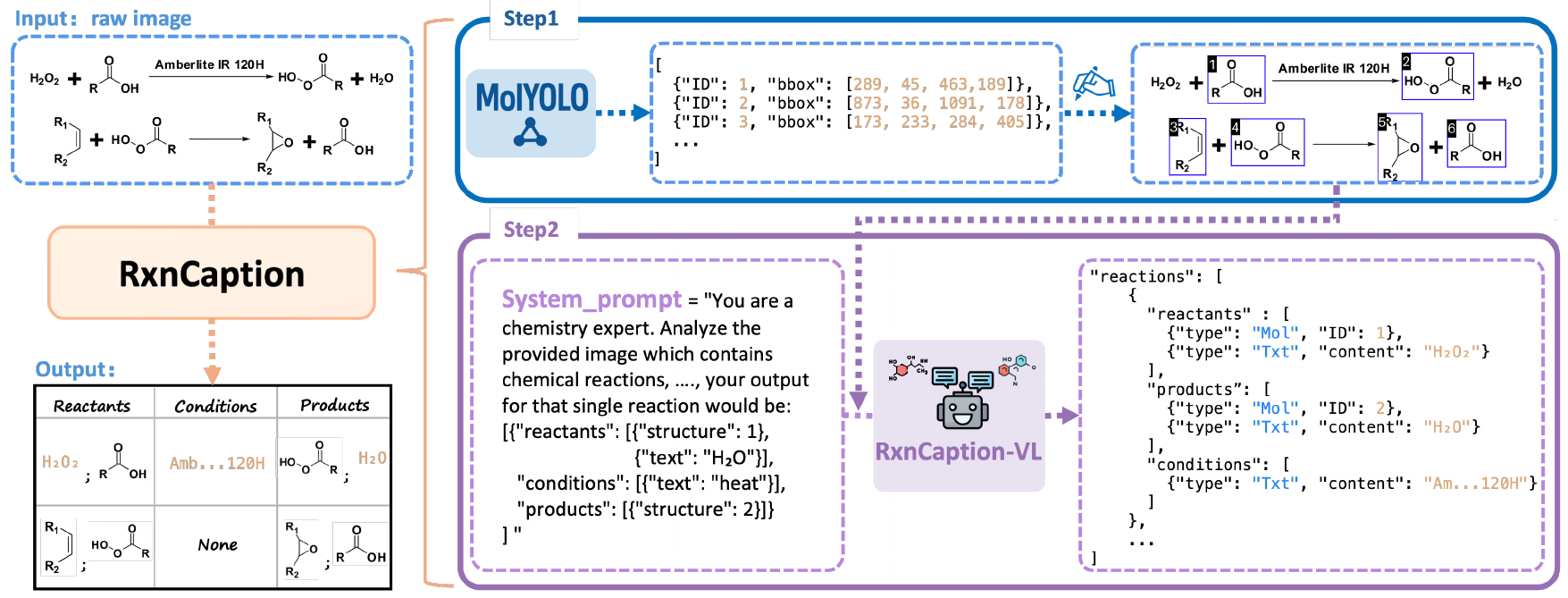}
    \caption{The RxnCaption framework reformulates the chemical Reaction Diagram Parsing (RxnDP) into an image captioning task. Using the ``BBox and Index as Visual Prompt" (BIVP) strategy, it employs our advanced molecular detector, MolYOLO, to pre-annotate input images with molecular bounding boxes and indices. This enables LVLMs to complete the task by describing the reaction in natural language, which fully leveraging their inherent capabilities.}
    \label{fig:overview_RxnCaptionVL}
\end{figure*}

Chemical reaction diagrams are the most commonly used representation of reactions in papers due to their simplicity, intuitiveness, and strong expressive capability. However, they are in the non-machine-readable format. As shown in Figure ~\ref{fig:overview_RxnCaptionVL}, the Reaction Diagram Parsing (RxnDP) task takes chemical reaction diagrams as input, with the goal of outputting all chemical reactions. Each reaction includes three roles: reactants, conditions, and products. From the perspective of human reading and understanding, this task can naturally be broken down into three steps: 1) Molecular structure detection (outputting bounding boxes); 2) Reaction extraction (assembling molecular bboxes and text conditions into complete reactions and assigning roles); 3) Post-processing (using OCSR models to recognize molecular bboxes as SMILES structures and OCR to recognize text conditions).

Recent studies~\cite{wilary2023reactiondataextractor,qian2023rxnscribe,RxnIm} have applied deep learning to RxnDP. RxnScribe~\cite{qian2023rxnscribe} uses the Pix2Seq~\cite{chen2021pix2seq} model for simultaneous molecule detection and reaction extraction via a sequence generation approach, termed as the ``Bbox and Role in One Step" (BROS) strategy. RxnIM~\cite{RxnIm} was the first to apply the LVLM to RxnDP, also using the BROS strategy and significantly increasing training data through data synthesis. However, its performance improvement over RxnScribe is modest, and it struggles with out-of-distribution generalization. Thus, the application of LVLMs in this field has not yet led to the breakthroughs seen in other domains.

To achieve significant progress in RxnDP with LVLMs, improvements are needed in two areas, addressing RxnIM's limitations.
Firstly, the BROS strategy may not suit LVLMs, as coordinate prediction falls outside their core competencies~\cite{li2025lmm, kang2025your}. Instead, prediction strategies should align with LVLMs' inherent strengths. We propose the RxnCaption framework, which allows LVLMs to parse reactions by describing images, leveraging their natural language generation capabilities. Using a visual prompt approach~\cite{wu2024visual}, we employ MolYOLO, a specialized molecular detection model we developed, to pre-draw molecular bounding boxes and indice on the image, creating a pre-annotated image. The LVLM then parses the reaction by describing the image, referencing molecular indices, and using OCR for text recognition. This ``BBox and Index as Visual Prompt" (BIVP) strategy enables parsing entirely in natural language, maximizing their potential.
Secondly, from a data perspective, existing RxnDP datasets are limited: the \texttt{RxnScribe-train} set contains only a few samples from real papers, while the \texttt{RxnIM-train} set, though larger, is synthetic and differs from real images, hindering model generalization. To address this, we created the \texttt{RxnCaption-15k} dataset, a large, high-quality, diverse RxnDP dataset. 

Using the RxnCaption framework and \texttt{RxnCaption-15k} dataset, we developed the RxnCaption-VL model, achieving state-of-the-art performance on both \texttt{RxnScribe-test} and \texttt{RxnCaption-15k-test} datasets. We hope that our methods, data, and models will advance RxnDP task and support the development of AI for chemistry.

In summary, our main contributions are:

1. We introduced the RxnCaption method, which transforms chemical reaction diagram parsing into an image description task using visual prompts, bypassing the coordinate prediction challenge that LVLMs struggle with, thus fully leveraging LVLM's inherent capabilities.

2. We created \texttt{RxnCaption-15k}, the largest dataset for chemical reaction diagram parsing from real papers, an order of magnitude larger than the \texttt{RxnScribe} dataset.

3. We developed MolYOLO, a state-of-the-art molecular structure detector, and integrated it into the RxnCaption framework for precise visual prompts.

4. Using these methods, data, and models, we trained RxnCaption-VL, the leading LVLM model for chemical reaction diagram parsing.

\section{Pilot Study and Our Insights}
\label{sec:3}

\begin{figure*}
    \centering
    \includegraphics[width=1.0\linewidth]{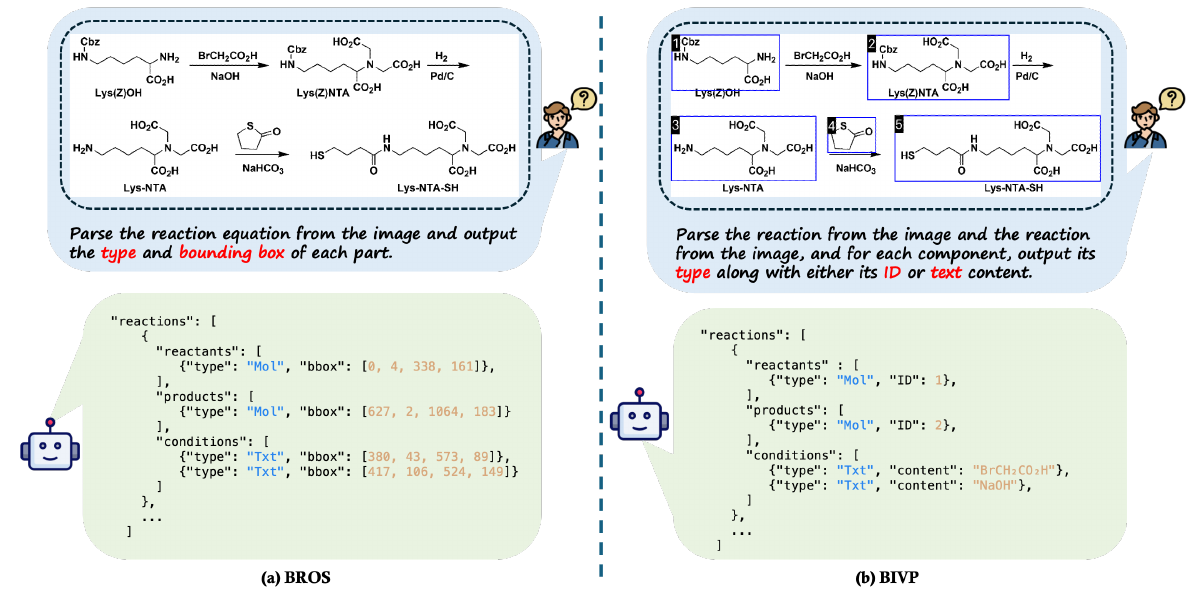}
    \caption{Comparison of ``Bbox and Role in One Step'' (BROS) and ``BBox and Index as Visual Prompt'' (BIVP) strategies.}
    \label{fig:BROS_vs_BIVP__v1}
\end{figure*}

\subsection{Definition of Reaction Diagram Parsing Task}
\label{sec:3-1}

In this section, we define the Reaction Diagram Parsing (RxnDP) task as \(
R = \mathcal{F}(I)
\),
where \( I \) represents the input chemical reaction diagram, \( \mathcal{F} \) is the RxnDP model, and \(\mathcal{R} = \{ R_i \}\) is the set of all chemical reactions in the diagram.

Each reaction \(R_i\) consists of three roles:  
\[
R_i=\bigl(\underset{\text{reactants}}{\mathcal C_{\text{react}}},\;
\underset{\text{conditions}}{\mathcal C_{\text{cond}}},\;
\underset{\text{products}}{\mathcal C_{\text{prod}}}\bigr)
\]

Each role \(\mathcal{C}_{\text{role}} = \{ c_j \}\) (\(\text{role} \in \{\text{react}, \text{cond}, \text{prod}\}\)) is composed of several components. The components can be in two modalities: molecular structure diagrams and text. A molecular structure diagram component (referred to as molecular component) is represented by the bounding box coordinates of the molecule in the diagram, while a text component is represented directly by its textual content.

% \begin{table}[htbp]
%     \centering
%     \renewcommand{\arraystretch}{1.2}
%     % \resizebox{0.45\textwidth}{!}{%
%     \resizebox{\columnwidth}{!}{%
%         \begin{tabular}{lccc}
%             \hline
%             \textbf{Task} & \textbf{metric} & \textbf{Gemini} & \textbf{GPT4o} & \textbf{QwenVL} \\
%             \hline
%             \textbf{BROS f1 (Soft Match)} & 35.4 & 0.3 & 4.4 \\
%             \textbf{VQA acc (avg)} & 75.91 & 61.59 & 73.63 \\
%             \textbf{BIVP f1 (Soft match)} &81.04  &57.57  &66.47  \\
%             \hline
%         \end{tabular}%
%     }
%     \caption{Results of different VLMs across different tasks.}
%     \label{tab:vlm_eval_bios_vqa_bivp}
% \end{table}

% \begin{table}[htbp]
%     \centering
%     % \renewcommand{\arraystretch}{1.2}
%     \resizebox{\columnwidth}{!}{%
%         \begin{tabular}{lccc}
%             \noalign{\hrule height 1.2pt}  % 顶部粗线
%             \textbf{Task} & \textbf{metric} & \textbf{Gemini} & \textbf{GPT4o} & \textbf{QwenVL} \\
%             \hline
%             \textbf{BROS} & F1 & 35.4 & 0.3 & 4.4 \\
%             \textbf{VQA} & Acc. & 75.91 & 61.59 & 73.63 \\
%             \textbf{BIVP} & F1 & 81.04 & 57.57 & 66.47 \\
%             \noalign{\hrule height 1.2pt}  % 底部粗线
%         \end{tabular}%
%     }
%     \caption{Results of LVLMs on the pilot studies. We use SoftMatch metric for RxnDP task.}
%     \label{tab:vlm_eval_bios_vqa_bivp}
% \end{table}

\begin{table}[htbp]
    \centering
    \begin{tabular}{lcccc}
        \toprule
        \textbf{Task} & \textbf{Metric} & \textbf{Gemini} & \textbf{GPT4o} & \textbf{QwenVL} \\
        \midrule
        \textbf{BROS} & F1 & 35.4 & 0.3 & 4.4 \\
        \textbf{VQA} & Acc. & 75.9 & 61.6 & 73.6 \\
        \textbf{BIVP} & F1 & 81.0 & 57.6 & 66.6 \\
        \bottomrule
    \end{tabular}
    \caption{Results of LVLMs on the pilot studies. We use SoftMatch metric for RxnDP task.}
    \label{tab:vlm_eval_bios_vqa_bivp}
\end{table}

\subsection{Reflection on the BROS Strategy}
\label{sec:3-2}

Initially, we evaluated the RxnDP performance of leading LVLMs using the approach from RxnScribe~\cite{qian2023rxnscribe} and RxnIm~\cite{RxnIm}. These models were tasked with predicting both the bounding boxes (bboxes) of molecular components and their roles simultaneously, in a method we term "Bbox and Role in One Step" (BROS).

We tested models such as GPT-4o~\cite{hurst2024gpt}, Gemini-2.5-Pro~\cite{team2023gemini}, and Qwen-VL-max~\cite{bai2023qwentechnicalreport} in a zero-shot scenario on a subset of the \texttt{RxnScribe-test} set, labeled \texttt{RxnScribe-test-slct}.\footnote{This subset includes only samples where all molecules participate in the reaction and have complete bbox annotations, excluding interference from free molecules.} Details on the prompt templates and evaluation metrics are in Appendix B and \S ~\ref{sec:experiment_setup}. As shown in Table~\ref{tab:vlm_eval_bios_vqa_bivp}, despite their strong performance in general tasks, these models perform poorly on RxnDP using the BROS strategy.

This is unexpected, given the extensive interleaved text and image data from chemical literature and web sources in LVLM pre-training datasets~\cite{li2024omnicorpus}. To delve deeper, we devised a visual question answering (VQA) task to assess understanding of reaction diagrams, with questions like \textit{“How many chemical reactions are depicted?”} and \textit{“Does the diagram include cyclic structures?”} (details in Appendix B). As shown in Table~\ref{tab:vlm_eval_bios_vqa_bivp}, mainstream LVLMs exhibit a good understanding of reaction diagrams, with Gemini-2.5 Pro achieving an accuracy of 75.9\%.

Comparative analysis indicates that the core issue is the LVLMs' need to predict molecular bboxes.
The BROS strategy lies beyond LVLMs' inherent strengths~\cite{li2025lmm, kang2025your}. RxnIm encountered similar challenges with BROS, addressing them through multi-stage training, initially using synthetic data to enhance molecular bbox prediction. However, this approach is complex and costly, with limited performance gains and a lack of generalization capability (see \S~\ref{sec:main_results}).

\subsection{Our Insights and Preliminary Verification}
\label{sec:3-3}

The VQA results confirm that mainstream VLMs can effectively understand chemical reaction diagrams, indicating they possess relevant domain knowledge and capabilities. This leads to a key insight: instead of forcing LVLMs to learn new skills during fine-tuning, we should develop prediction strategies that leverage their inherent capabilities. Inspired by research on visual prompts~\cite{wu2024visual}, we introduce the \textbf{BBox and Index as Visual Prompt (BIVP)} strategy:

(1) Pre-annotate molecular component bounding boxes on the image.

(2) Add index numbers next to the bounding boxes.

This approach bypasses the LVLM's limitations in bbox prediction, allowing it to focus on reaction parsing. To validate this idea, we applied the BIVP strategy to the \texttt{RxnScribe-test-slct} dataset. For ease of validation, we utilized the ground truth molecular component bounding boxes from the dataset. As shown in Table~\ref{tab:vlm_eval_bios_vqa_bivp}, the performance of mainstream VLMs significantly improved, confirming the effectiveness of the BIVP strategy and the validity of our previous analysis.

\section{Method}

\begin{figure*}[t]
    \centering
    \vspace{-5pt}
    \includegraphics[width=\textwidth]{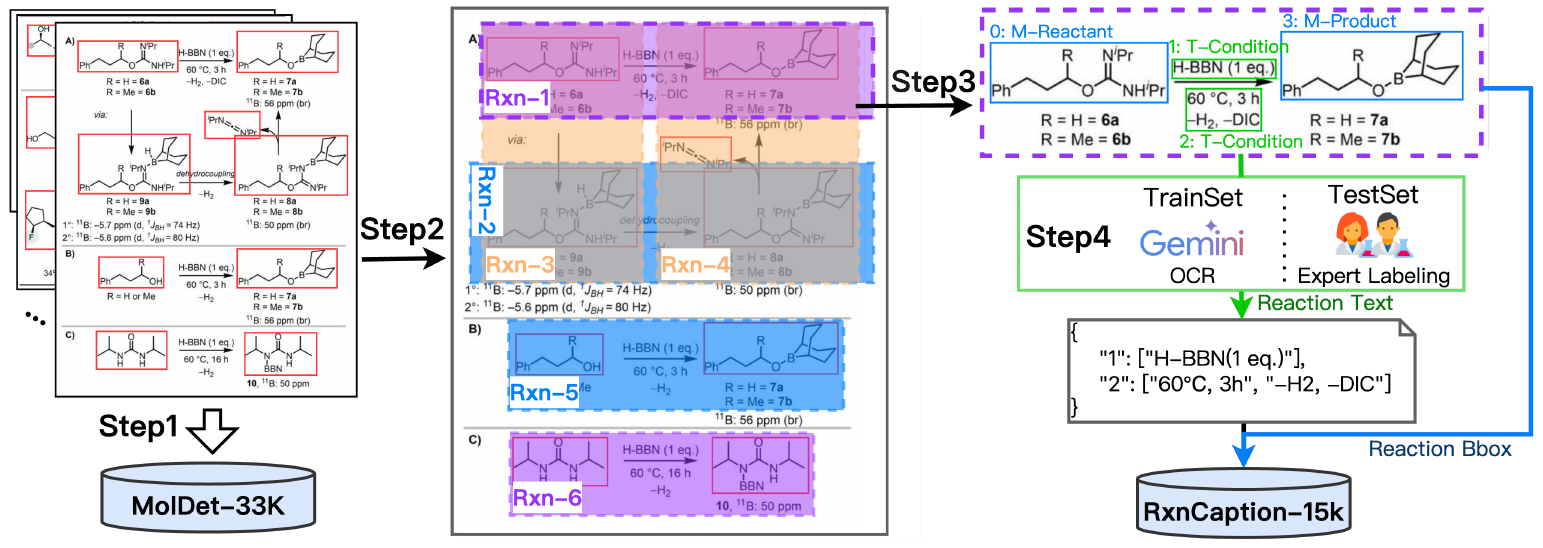}
    \caption{Construction pipeline of \texttt{RxnCaption-15k} dataset.}
    \label{fig:rxncaption-15k_construction}
\end{figure*}

\subsection{MolYOLO}
\label{sec:4-1}

The BIVP strategy, which involves pre-drawing bounding boxes (bboxes) and indices on chemical reaction diagrams, significantly enhances LVLM performance in parsing reactions. This requires a robust molecular detector. Existing models and VLMs perform poorly in this task (see Table~\ref{tab:molyolo_performance}), prompting us to develop a high-precision detector.

\subsubsection{\texttt{MolDet-33k} dataset}

We created the \texttt{MolDet-33k} dataset to train MolYOLO. This dataset includes images from Page Scope and Fig/Tab Scope, sourced from about 3000 organic chemistry papers. PDFs were converted to JPGs for Page Scope images. Using DocLayoutYOLO, we extracted Fig and Tab bboxes to crop Fig/Tab Scope images. Professional annotators labeled 219,721 molecular bboxes, mapped to Fig/Tab Scope images. We filtered out images without bboxes, resulting in 12,209 Page Scope and 21,155 Fig/Tab Scope images. Data from papers published by June 2024 formed the training set, while those from July 2024 were for validation.

\subsubsection{Training and Performance}
\begin{table}[htbp]
\centering
\begin{tabular}{lcc}
\toprule
Model & P IoU@0.5 & R IoU@0.5 \\
\midrule
MolDetect & 0.84 & 0.77 \\
YoDe      & 0.89 & 0.75 \\
MolYOLO (ours) & \textbf{0.98} & \textbf{0.98} \\
\bottomrule
\end{tabular}
\caption{Performance on \texttt{MolDet-33k-test} set.}
\label{tab:molyolo_performance}
\end{table}
We based our detector on the YOLOv10~\cite{wang2024yolov10} architecture, using the \texttt{MolDet-33k-train} set and the \texttt{YoDe} dataset~\cite{zhou2023yode} (1,738 images, 15,887 bboxes). We evaluated MolYOLO, YoDe, MolDetect\footnote{\url{https://github.com/Ozymandias314/MolDetect}} on \texttt{MolDet-33k-test}. As shown in Table~\ref{tab:molyolo_performance}, MolYOLO achieved state-of-the-art performance, with particularly high accuracy and recall, demonstrating its strong detection capabilities.

\begin{figure}[h]
    \centering
    \vspace{-5pt}
    % --- 左侧子图 ---
    \begin{subfigure}[t]{0.48\linewidth}
        \centering
        \includegraphics[height=1in]{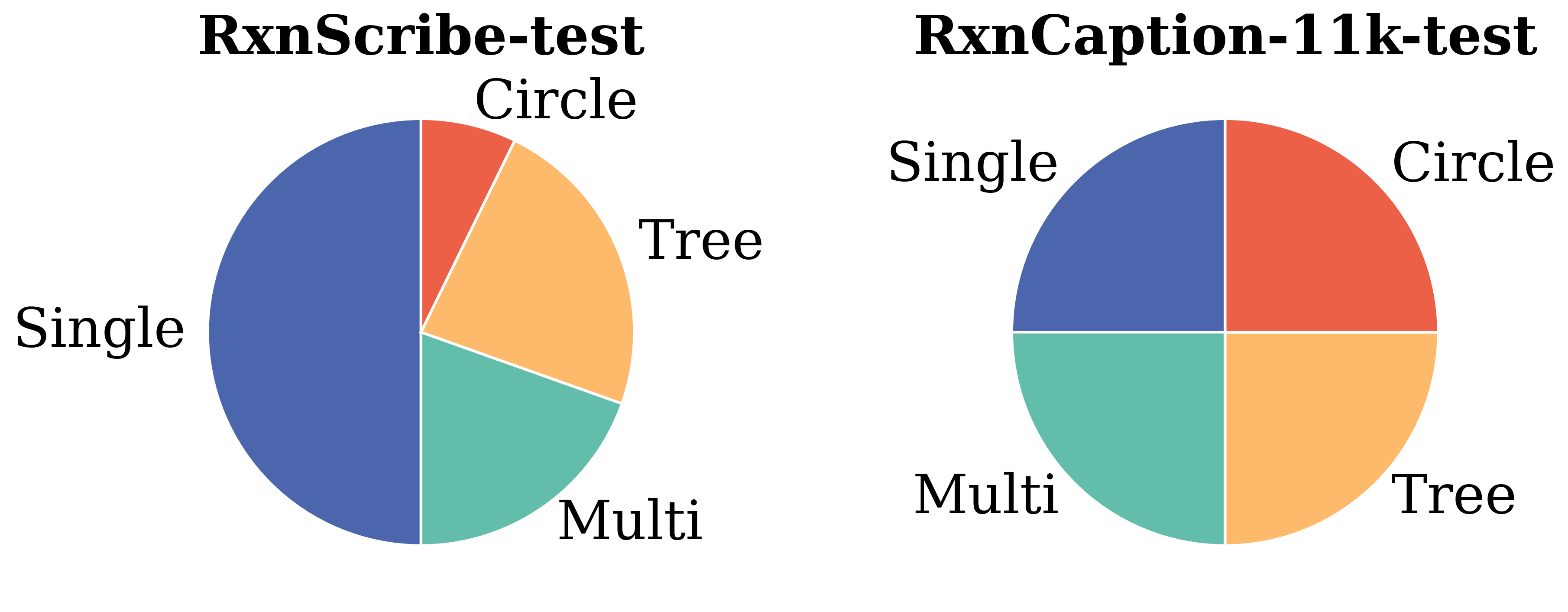}
        \caption{Layout distribution of \texttt{RxnScribe-test} and \texttt{RxnCaption-15k-test}.}
        \label{fig:RxnDP_testset_dist}

    \end{subfigure}
    \hfill
    % --- 右侧子图 ---
    \begin{subfigure}[t]{0.48\linewidth}
        \centering
        \includegraphics[width=\linewidth]{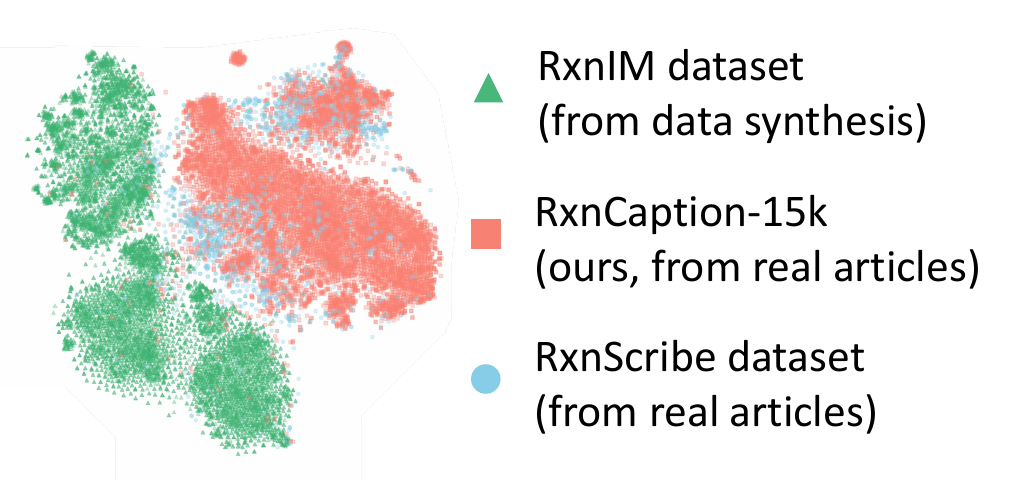}
        \caption{t-SNE visualization of the image embeddings for \texttt{RxnScribe}, \texttt{RxnIm}, and \texttt{RxnCaption-15k}.}
        \label{fig:t_SNE}
    \end{subfigure}
    \vspace{-6pt}
    \caption{(a)~t-SNE visualization of dataset embeddings; (b)~Layout distribution comparison between \texttt{RxnScribe} and \texttt{RxnCaption-15k}.}
    \label{fig:tSNE_layout}
\end{figure}

\subsection{\texttt{RxnCaption-15k} Dataset}
\label{sec:4-2}

Existing RxnDP datasets have notable limitations: \texttt{RxnScribe} dataset, derived from real papers, includes only 1378 samples, while \texttt{RxnIM} dataset, though large, is artificially synthesized, limiting diversity and model generalization (see \S~\ref{sec:main_results}). To overcome these issues, we developed the \texttt{RxnCaption-15k} dataset—a large, high-quality, and diverse RxnDP dataset.

\subsubsection{Construction}

\texttt{RxnCaption-15k} uses the same chemical paper sources and Fig/Tab-level images as \texttt{MolDet-33k}. Its construction involves four steps (Figure ~\ref{fig:rxncaption-15k_construction}):
\textbf{(1) Molecular Structure Annotation}: Annotate molecular structure bboxes, completed during \texttt{MolDet-33k} construction.
\textbf{(2) Reaction Region Annotation}: Use irregular polygons to mark each chemical reaction area.
\textbf{(3) Component Role Annotation}: Annotate the chemical role of each reaction component, with text components following RxnScribe guidelines.
\textbf{(4) Text Content Extraction}:
For training Set, we use Gemini-2.5 Pro's OCR for automated text extraction (details in Appendix C). For validation set, we use manual annotation for accuracy.

\subsubsection{Statistics and Comparison}
\begin{table}[h]
\centering
\setlength{\tabcolsep}{3pt}
\begin{tabular}{l>{\centering\arraybackslash}m{1.8cm}cc}  
\toprule
\multicolumn{1}{c}{\multirow{2}{*}{Dataset}} & \multicolumn{1}{c}{\multirow{2}{*}{Source}} & \multicolumn{2}{c}{Data Statistics} \\
\cmidrule(lr){3-4}
 &  & Train & Test \\
\midrule
\texttt{RxnIm} & synthesis & 
\#Img: 48,160 & 
\#Img: 6,020\\
\specialrule{0.5pt}{1pt}{1pt}
\texttt{RxnScribe} & real & 
\begin{tabular}[c]{@{}c@{}}\#Img: 1,240 \\ \#Rxn: 3,384\end{tabular} & 
\begin{tabular}[c]{@{}c@{}}\#Img: 138 \\ \#Rxn: 392\end{tabular} \\
\specialrule{0.5pt}{1pt}{1pt}
ours & real & 
\begin{tabular}[c]{@{}c@{}}\#Img: \textbf{10,112} \\ \#Rxn: \textbf{24,716}\end{tabular} & 
\begin{tabular}[c]{@{}c@{}}\#Img: \textbf{400} \\ \#Rxn: \textbf{2,829}\end{tabular} \\
\bottomrule
\end{tabular}
\caption{Statistics of the RxnDP datasets. `ours' means \texttt{RxnCaption-15k}.}
\label{tab:RxnDP_dataset}
\end{table}
Following \texttt{MolDet-33k}'s division strategy, samples from papers published by June 2024 were used for training, and those from July 2024 for validation. Chemical reaction diagrams are categorized into Single-line, Multi-line, Tree, and Cyclic. Figure~\ref{fig:RxnDP_testset_dist} shows that the \texttt{RxnScribe-test} set is imbalanced, with the Single-line category dominating at 50\%, while Tree/Cyclic categories are only 23.2\%/7.2\%. To address this, \texttt{RxnCaption-15k-test} set was carefully selected to ensure category diversity, with 100 Fig/Tab images from each category. The final dataset statistics are summarized in Table ~\ref{tab:RxnDP_dataset}.

To evaluate the distribution of \texttt{RxnCaption-15k}, \texttt{RxnScribe}, and \texttt{RxnIM} datasets, we used the CLIP visual encoder~\cite{radford2021learning} for feature extraction and visualized their distribution with t-SNE~\cite{maaten2008visualizing} (see Figure ~\ref{fig:t_SNE}). The results indicate:
\textbf{(1) Real Data Diversity} \texttt{RxnCaption-15k} and \texttt{RxnScribe} datasets, both from real papers, show significant feature overlap. However, \texttt{RxnCaption-15k}'s broader coverage confirms its advantage in quantity and diversity.
\textbf{(2) Synthetic Data Domain Shift} \texttt{RxnIM} dataset's distribution shows little overlap with real datasets, highlighting a domain gap. This explains RxnIM's performance bottleneck: despite large data size, its performance improves litte on the \texttt{RxnScribe-test} set, and even performs worse on \texttt{RxnCaption-15k} compared to RxnScribe (see \S~\ref{sec:main_results}).

\subsection{RxnCaption-VL}
\label{sec:4-3}

Building on our pilot study and insights from \S \ref{sec:3}, we introduce the RxnCaption framework, illustrated in Figure~\ref{fig:overview_RxnCaptionVL}. The framework leverages LVLMs to generate image descriptions, enabling the parsing of chemical reaction diagrams. As detailed in \S 3.1, chemical reactions involve molecular and text components. For text component, the LVLM is tasked to directly extract the textual content and assign the role. For molecular component, we apply the BIVP strategy from \S 3.3. This involves pre-annotating images with bounding boxes and index numbers, allowing LVLMs to reference these indices to output information about the chemical reaction and the role of each molecule.
For training, we use ground truth bounding boxes for molecular component within reactions and MolYOLO-detected boxes for those outside. During inference, we rely solely on MolYOLO-detected bounding boxes.
We refer to images with these molecular bounding boxes and index markings as pre-annotated images. These are input into the LVLM during training and inference, requiring the model to output all chemical reaction information in JSON format, as shown in Figure~\ref{fig:overview_RxnCaptionVL}.

We expanded the \texttt{RxnScribe-train} set from 1,240 to 3,720 images to balance the ratio of and combined it with the \texttt{RxnCaption}\texttt{-15k-train} set of 15,128 images. We additionally applied double augmentation to images containing reversible reactions, right-to-left reactions, and bottom-to-top reactions, resulting in a training set of 23,432 images. Using this, we fine-tuned the Qwen2.5-VL-7B, resulting our RxnCaption-VL model.
\section{Experiments}

\begin{table*}[h]
\centering
\caption{Model performance comparison on the \texttt{RxnScribe-test} and \texttt{RxnCaption-15k-test} datasets. ``RxnScribe\_official'' refers to the model from ~\cite{qian2023rxnscribe}. ``RxnScribe\_\texttt{w/15k}'' is the RxnScribe model retrained with our dataset. Best scores are in \textbf{bold}, second best are \underline{underlined}.}
\label{tab:model_performance}
\setlength{\tabcolsep}{6pt}
\renewcommand{\arraystretch}{1}
\renewcommand{\heavyrulewidth}{1.2pt}
\resizebox{\textwidth}{!}{
\begin{tabular}{@{}c c c *{6}{c} *{6}{c}@{}}
\toprule
\multirow{4}{*}{\bfseries Model} & 
\multirow{4}{*}{\bfseries Strategy} &
\multicolumn{6}{c}{\bfseries \texttt{RxnScribe-test}} & 
\multicolumn{6}{c}{\bfseries \texttt{RxnCaption-15k-test}} \\
\cmidrule(lr){3-8} \cmidrule(lr){9-14}
 &  & \multicolumn{3}{c}{\bfseries Hybrid Match} & \multicolumn{3}{c}{\bfseries Soft Match} & \multicolumn{3}{c}{\bfseries Hybrid Match} & \multicolumn{3}{c}{\bfseries Soft Match} \\
\cmidrule(lr){3-5} \cmidrule(lr){6-8} \cmidrule(lr){9-11} \cmidrule(lr){12-14}
 &  & \textbf{P} & \textbf{R} & \textbf{F1} & \textbf{P} & \textbf{R} & \textbf{F1} & \textbf{P} & \textbf{R} & \textbf{F1} & \textbf{P} & \textbf{R} & \textbf{F1} \\
\midrule
\multicolumn{14}{c}{\textbf{Trained Model}} \\
\midrule
\multirow{2}{*}{\textbf{RxnCaption-VL}} 
 & BIVP   & 71.6 & \textbf{72.7} & \textbf{72.2} & \textbf{85.3} & \textbf{87.1} & \textbf{86.2} &  \underline{60.3} & \textbf{59.3} & \textbf{59.8} &  \underline{71.3} & \textbf{69.4} & \textbf{70.4} \\
 & BROS   & 69.6 & 68.9 & 69.2 & 76.2 & 76.2 & 76.2 & 57.0 & \underline{57.5} & \underline{57.2} & 66.4 & 67.4 & 66.9 \\
\textbf{RxnScribe\_\texttt{w/15k}} & BROS & \textbf{72.4} & 69.1 & \underline{70.7} & \underline{84.1} & 81.7 & \underline{82.8} & \textbf{61.2} & 38.7 & 47.4 & \textbf{72.1} & 44.7 & 55.2 \\
\textbf{RxnScribe\_official} & BROS & \underline{72.3} & 66.2 & 69.1 & 83.8 & 76.5 & 80.0 & 47.4 & 27.6 & 34.9 & 62.1 & 36.4 & 45.9 \\
\textbf{RxnIM} & BROS & 71.0 & \underline{70.1} & 70.5 & 79.2 & 74.7 & 76.9 & 48.8 & 30.3 & 37.4 & 52.9 & 32.8 & 40.5 \\
\midrule
\multicolumn{14}{c}{\textbf{Open-source Model}} \\
\midrule
\multirow{2}{*}{\textbf{Intern-VL3-78B}} 
 & BIVP  & 33.8 & 44.1 & 38.3 & 45.8 & 59.8 & 51.9 & 13.0 & 15.5 & 14.1 & 26.6 & 32.0 & 29.0 \\
 & BROS  & 0.0  & 0.0  & 0.0  & 0.0  & 0.0  & 0.0  & 0.1  & 0.1  & 0.1  & 0.3  & 0.2  & 0.3  \\
\multirow{2}{*}{\textbf{Qwen2.5-VL-7B}} 
 & BIVP  & 6.0  & 4.1  & 4.9  & 55.8 & 36.0 & 43.8 & 2.9  & 0.9  & 1.4  & 33.0 & 10.3 & 15.6 \\
 & BROS  & 0.0  & 0.0  & 0.0  & 0.0  & 0.0  & 0.0  & 0.0  & 0.0  & 0.0  & 0.0  & 0.0  & 0.0  \\
\multirow{2}{*}{\textbf{Qwen2.5-VL-72B}} 
 & BIVP  & 51.9 & 48.5 & 50.1 & 70.0 & 66.3 & 68.1 & 30.9 & 23.8 & 26.9 & 52.8 & 41.6 & 46.5 \\
 & BROS  & 2.0  & 1.4  & 1.6  & 15.2 & 11.2 & 12.9 & 0.3  & 0.3  & 0.3  & 4.2  & 1.9  & 2.6  \\
\midrule
\multicolumn{14}{c}{\textbf{Closed-source Model}} \\
\midrule
\multirow{2}{*}{\textbf{Gemini-2.5-Pro}} 
 & BIVP  & 44.7 & 56.1 & 49.8 & 67.9 & \underline{86.5} & 76.1 & 38.9 & 42.1 & 40.4 & 64.2 & \underline{69.2} & \underline{66.6} \\
 & BROS  & 0.0  & 0.0  & 0.0  & 25.2 & 23.5 & 24.3 & 0.3  & 0.2  & 0.3  & 8.9  & 4.6  & 6.0  \\
\multirow{2}{*}{\textbf{GPT4o-2024-11-20}} 
 & BIVP  & 26.8 & 33.2 & 29.6 & 49.1 & 58.0 & 53.2 & 16.1 & 16.6 & 16.3 & 32.6 & 32.7 & 32.6 \\
 & BROS  & 0.3  & 0.3  & 0.3  & 2.0  & 1.8  & 1.9  & 0.0  & 0.0  & 0.0  & 0.4  & 0.3  & 0.3  \\
\multirow{2}{*}{\textbf{Qwen-VL-Max}} 
 & BIVP  & 50.0 & 46.9 & 48.4 & 71.1 & 67.6 & 69.3 & 34.0 & 29.0 & 31.3 & 55.3 & 48.4 & 51.6 \\
 & BROS  & 0.3  & 0.3  & 0.3  & 7.2  & 5.9  & 6.5 & 0.2  & 0.1  & 0.2  & 3.2  & 2.6  & 2.8  \\
\bottomrule
\end{tabular}
}
\end{table*}

\subsection{Experiment Setup}

\label{sec:experiment_setup}

\subsubsection{Implementation}

We train the MolYOLO model using the joint training set mentioned above in a single NVIDIA A100 GPU, initializing with YOLOv10-M pretrained weights. Optimization employs SGD (momentum=0.937, weight decay=\(5 \times 10^{-4}\)) over 30 epochs with a constant learning rate of 0.01. Input images are processed at 1024×1024 resolution with batch size 32, utilizing RandAugment for data augmentation and Automatic Mixed Precision (AMP) for accelerated training. All other hyperparameters are consistent with the YOLOv10 configuration.

The training of RxnCaption-VL was conducted using 8 NVIDIA A100 GPUs and the AdamW optimizer, with a maximum learning rate of \(1 \times 10^{-5}\). A cosine decay schedule was applied to the learning rate, along with a linear warm-up during the initial 5\% of iterations to ensure stable training. Each GPU processed a batch size of 1, and we employed gradient accumulation over 16 steps, leading to an effective batch size of 128. All model parameters were updated throughout the training process. We used DeepSpeed ZeRO Stage 2 to improve efficiency and reduce memory usage, enabling the effective training of large multimodal models. The model was trained for 10 epochs, optimizing the balance between performance and computational resources.

\subsubsection{Baselines}

We select two types of baseline models: the first type uses current mainstream general VLMs under zero-shot setting, including GPT4o, Gemini-2.5-Pro (all following metrics were tested in October 2025), and Qwen2.5-VL and Intern-VL3; the second type is the specialized models for RxnDP task, primarily RxnScribe and RxnIM.

During testing, we compare the BROS and BIVP strategies for the general VLMs, while the specialized models continue to utilize their native BROS strategy.

\subsubsection{Metric}

We follow the evaluation framework of RxnScribe~\cite{qian2023rxnscribe}, constructing an assessment system at the chemical reaction granularity. Given the predicted reaction set \(\mathcal{R}^{\text{pred}}=\{R_i^{\text{pred}}\}\) and the ground truth reaction set \(\mathcal{R}^{\text{gt}}=\{R_i^{\text{gt}}\}\), performance metrics are calculated through reaction instance matching: a match between instances in the predicted and true sets is recorded as a true positive (TP), unmatched predicted reactions are false positives (FP), and unmatched true reactions are false negatives (FN).

To evaluate model performance, we employ two complementary strategies: \textbf{SoftMatch} and \textbf{HybridMatch}.
SoftMatch follows the RxnScribe protocol by excluding text components and merging condition components into reactants. A reaction is counted as a TP only if all molecular bounding boxes in reactants and products achieve an IoU $\geq$ 0.5. This applies to both BROS and BIVP.

HybridMatch adapts to differences in output formats across methods. For BROS, it uses the strict HardMatch rule from RxnScribe, requiring all molecular components (reactants, products, and conditions) to reach IoU $\geq$ 0.5. If text components fail IoU matching, a textual matching strategy is applied (will be described as follows).
For BIVP, molecular components must have exact bounding-box matches (IoU $\geq$ 0.5). Text within \textbf{reactants} and \textbf{products} must match exactly, while \textbf{condition} text is considered correct if its normalized edit distance $\leq$ 0.2. Overall, BIVP is evaluated more strictly, since BROS considers text matched if either IoU or text content matches.

Finally, based on the TP/FP/FN statistics, we compute the precision, recall, and F1 scores under both SoftMatch and HybridMatch.

\subsection{Main Results}

\label{sec:main_results}

We compared our RxnCaption-VL model with the baselines mentioned in \S~\ref{sec:experiment_setup} on the \texttt{RxnScribe-test} and \texttt{RxnCaption-15k-test}, as shown in Table~\ref{tab:model_performance}. It is evident that on both datasets, our RxnCaption-VL (using BIVP) achieved the best F1 scores across both HybridMatch and SoftMatch metrics. Notably, on the more challenging \texttt{RxnCaption-15k-test} set, RxnCaption-VL outperformed the strongest competing model, Gemini-2.5-Pro (BIVP), by \textbf{19.4} percentage points in HybridMatch F1 and \textbf{3.8} percentage points in SoftMatch F1, demonstrating its superior performance.
Additionally, we observed that Gemini-2.5-Pro exhibits excellent performance on the RxnDP task. When using the BIVP strategy, it secured the highest F1 scores among all \textbf{general} VLMs.

\subsubsection{BROS vs BIVP}

As shown in Table~\ref{tab:model_performance}, the BIVP strategy demonstrated a comprehensive advantage over the BROS strategy. BIVP consistently outperformed BROS across all general LVLMs. We further compared the training of RxnCaption-VL using both BROS and BIVP strategies on the same training data. The results showed that on the \texttt{RxnScribe-test} set, the BIVP strategy improved the Hybrid-F1 by \textbf{3.0} percentage points and the Soft-F1 by a remarkable \textbf{10.0} percentage points compared to BROS. These results convincingly demonstrate the effectiveness of our proposed BIVP strategy for the RxnDP task.

\subsubsection{Real Data vs Synthetic Data}

We further examined the differences between real paper data and synthetic data for the RxnDP task by comparing the performance of the RxnScribe, RxnCaption-VL, and RxnIM models. Despite the fact that the RxnIM training dataset (in terms of the number of images) is 38 times larger than that of RxnScribe and 4 times larger than that of RxnCaption-VL, its performance is noticeably inferior to these two models. Particularly on the \texttt{RxnCaption-15k-test} set, the SoftMatch F1 score for RxnIM was lower by \textbf{5.4} percentage points compared to RxnScribe\_official, and by a significant \textbf{29.9} percentage points compared to RxnCaption-VL. As discussed in \S~\ref{sec:4-2}, the t-SNE visualization reveals a significant domain shift between RxnIM's synthetic data and real data. These experimental conclusions further validate the limitations of current data synthesis methods.

\subsubsection{Training RxnScribe with \texttt{RxnCaption-15k}}

To validate the effectiveness of \texttt{RxnCaption-15k}, we retrained the RxnScribe model using the BROS strategy, while keeping the data configuration identical to that of RxnCaption-VL and following the official RxnScribe training scripts. This retrained model, referred to as RxnScribe\_\texttt{w/15k}, was compared with the official open-source weights (RxnScribe\_official), as shown in Table~\ref{tab:model_performance}. The RxnScribe\_\texttt{w/15k} showed significant improvement on the \texttt{RxnCaption-15k-test} set, with its Hybrid-F1 score increasing by \textbf{12.5} percentage points. However, on the original \texttt{RxnScribe-test} set, its Hybrid-F1 score showed a slight increase of 1.6 points, and the Soft-F1 score improved by 2.8 points. This indicates that while our dataset enhances performance on similar real-world data, the limited capacity of the Pix2Seq model may struggle to perfectly generalize across both data distributions. Moreover, constrained by the BROS strategy and model size, RxnScribe\_\texttt{w/1e5k} still lags behind RxnCaption-VL.

\subsection{Ablation Studies}

\subsubsection{Influence of Molecular Detector}
% \begin{table}[htbp]
%     \centering
%     \setlength{\tabcolsep}{6pt}
%     \renewcommand{\arraystretch}{1}
%     \begin{tabular}{c|c|cc|cc}
%         \toprule
%         \multirow{2}{*}{\textbf{LVLM}} & \multirow{2}{*}{\textbf{Detector}} & \multicolumn{2}{c|}{\textbf{RxnScribe-test}} & \multicolumn{2}{c}{\textbf{RC-15k-test}} \\
%         \cline{3-6}
%         & & \textbf{S-F1} & \textbf{H-F1} & \textbf{S-F1} & \textbf{H-F1} \\
%         \midrule
%         \multirow{3}{*}{\textbf{\textit{G}}} & \textbf{YoDe} & 54.5 & 34.7 & 59.4 & 36.0 \\
%         & \textbf{MolDetect} & 72.3 & 47.9 & 60.9 & 37.8 \\
%         & \textbf{Ours} & 76.1 & 49.8 & \underline{66.6} & 40.4 \\
%         \midrule
%         \multirow{3}{*}{\textbf{\textit{RC}}} & \textbf{YoDe} & 61.5 & 53.3 & 60.2 & 49.2 \\
%         & \textbf{MolDetect} & \underline{84.4} & \underline{71.8} & 64.0 & \underline{53.2} \\
%         & \textbf{Ours} & \textbf{88.1} & \textbf{75.5} & \textbf{67.6} & \textbf{55.5} \\
%         \bottomrule
%     \end{tabular}
%     \caption{Influence of Molecular Detector on BIVP strategy. \textbf{\textit{G}} means Gemini-2.5-Pro, \textbf{\textit{RC}} means our RxnCaption-VL. `RC-15k-test' means \texttt{RxnCaption-15k-test} dataset. `S-F1' and `H-F1' means F1 score on the SoftMatch and HybridMatch metrics. `Ours' means our MolYOLO.}
%     \label{tab:mol_det_influence_on_BIVP}
% \end{table}
\begin{table}[htbp]
    \centering
    \setlength{\tabcolsep}{6pt}
    \renewcommand{\arraystretch}{1}
    \begin{tabular}{c|c|cc|cc}
        \toprule
        \multirow{2}{*}{\textbf{LVLM}} & \multirow{2}{*}{\shortstack{\textbf{Molecular} \\ \textbf{Detector}}} & \multicolumn{2}{c|}{\textbf{\texttt{RxnScribe-test}}} & \multicolumn{2}{c}{\textbf{\texttt{RxnCaption-15k-test}}} \\
        \cline{3-6}
        & & \textbf{Soft-F1} & \textbf{Hybrid-F1} & \textbf{Soft-F1} & \textbf{Hybrid-F1} \\
        \midrule
        \multirow{3}{*}{\textbf{{Gemini}}} & \textbf{YoDe} & 54.5 & 34.7 & 59.4 & 36.0 \\
        & \textbf{MolDetect} & 72.3 & 47.9 & 60.9 & 37.8 \\
        & \textbf{MolYOLO} & 76.1 & 49.8 & \underline{66.6} & 40.4 \\
        \midrule
        \multirow{3}{*}{\textbf{{Ours}}} & \textbf{YoDe} & 61.5 & 53.3 & 60.2 & 49.2 \\
        & \textbf{MolDetect} & \underline{84.4} & \underline{71.8} & 64.0 & \underline{53.2} \\
        & \textbf{MolYOLO} & \textbf{88.1} & \textbf{75.5} & \textbf{67.6} & \textbf{55.5} \\
        \bottomrule
    \end{tabular}
    \caption{Influence of Molecular Detector on BIVP strategy. \textbf{Ours} means our RxnCaption-VL.}
    \label{tab:mol_det_influence_on_BIVP}
\end{table}
Accurate molecular detection is essential for the BIVP strategy's success. To validate our MolYOLO model, we substituted it with existing open-source detectors (YoDe, MolDetect) during the BIVP inference phase. As shown in Table~\ref{tab:mol_det_influence_on_BIVP}, the choice of detector significantly impacts overall performance for both our model and the Gemini baseline.

For our model, switching from the weakest detector, YoDe, to our proposed MolYOLO boosts the Hybrid-F1 score by a remarkable 18.9 points on the \texttt{RxnScribe-test} set (from 53.3 to 72.2). This trend holds for the Gemini baseline as well, which sees a 15.1-point improvement under the same conditions. Even when compared to the strong MolDetect baseline, our MolYOLO provides a consistent performance lift across all metrics. These results clearly demonstrate that a high-quality molecular detector is a critical prerequisite for achieving state-of-the-art performance with the BIVP strategy, validating the contribution of our specialized MolYOLO detector.

\begin{table}[!t]
\centering
\footnotesize  % 使用更小的字体
\setlength{\tabcolsep}{2pt}  % 进一步减少列间距
\renewcommand{\arraystretch}{0.9}  % 减少行高
\setlength{\heavyrulewidth}{1.2pt}  % 调整顶部和底部粗线的宽度

\caption{Error attribution analysis for the BIVP strategy. 
``MolYOLO" is the standard pipeline. ``GT bbox + MolYOLO" refers to a setup where ground truth boxes are supplemented by MolYOLO's non-overlapping predictions.
RS=\texttt{RxnScribe-test}, RC-15k=\texttt{RxnCaption-15k-test}.}
\label{tab:BIVP_two_stage_error_analysis}

\begin{tabular}{@{}l c cc cc@{}}
\toprule
\multirow{2}{*}{\makecell{\textbf{Detector} \\ \textbf{Setup}}} &  
\multirow{2}{*}{\textbf{Extractor}} &
\multicolumn{2}{c}{\textbf{RS}} &
\multicolumn{2}{c}{\textbf{RC-15k}} \\
\cmidrule(lr){3-4} \cmidrule(lr){5-6}
& & \textbf{Hybrid-F1} & \textbf{Soft-F1} & \textbf{Hybrid-F1} & \textbf{Soft-F1} \\
\midrule

\multirow{2}{*}{\textbf{MolYOLO}}
& \textbf{Gemini} & 49.8 & 76.1 & 40.4 & 66.2 \\
& \textbf{Ours}   & 72.2 & 86.2 & 59.8 & 70.4 \\
\midrule

\multirow{2}{*}{\makecell{\textbf{GT bbox + } \\ \textbf{MolYOLO}}}
& \textbf{Gemini}
& 51.5 (\textcolor{red}{+1.7})
& 82.2 (\textcolor{red}{+6.1})
& 46.4 (\textcolor{red}{+6.0})
& 73.5 (\textcolor{red}{+7.3}) \\
& \textbf{Ours}
& 73.4 (\textcolor{red}{+1.2})
& 87.6 (\textcolor{red}{+1.4})
& 63.8 (\textcolor{red}{+4.0})
& 76.3 (\textcolor{red}{+5.9}) \\
        \midrule
\textbf{MolYOLO} & \textbf{Ideal}
& 99.7 & 99.7 & 95.3 & 95.3 \\
\bottomrule
\end{tabular}
\end{table}

\subsubsection{Analysis of 2-Stage Strategy}
To quantify the performance loss from each component of our BIVP strategy, we conducted an error attribution analysis, with results shown in Table~\ref{tab:BIVP_two_stage_error_analysis}. We designed a specific ideal experiment, ``MolYOLO + GT bbox", to analyze the errors. In this setup, we started with the complete set of ground truth bounding boxes, ensuring perfect recall. Then, we added any bounding box predicted by MolYOLO that does not have a significant overlap (IoU $<$ 0.5) with any ground truth box. This setup effectively simulates a scenario with perfect recall but includes the noise from MolYOLO's false positive detections and some free molecules.

By comparing this ideal setup with the standard MolYOLO pipeline, we can isolate and analyze the impact of the detector itself, where the improvement gained from the ideal detector is highlighted in red. As shown in the table, the gain is relatively small, especially on the simpler \texttt{RxnScribe-test} dataset. Even on the more challenging \texttt{RxnCaption-15k-test} set, the reaction extraction model remains the dominant bottleneck. Furthermore, when we assume an ideal reaction extraction model in Stage 2 (as shown in the last row of the table), the performance ceiling imposed by MolYOLO is only reduced by \textbf{0.3} and \textbf{4.7} points, respectively. This further reinforces our conclusion that enhancing the LVLM’s reasoning capability is the most critical direction for future improvements.

Besides, since BIVP adopts a two-stage design, a natural concern is whether it introduces additional computational overhead compared with end-to-end approaches. We test RxnCaption-VL on a single A100 (80GB). BROS requires \textbf{16.88 s/img}, whereas BIVP achieves \textbf{9.65 s/img} (0.10 s for MolYOLO + 9.55 s for the LVLM). The speedup comes from BIVP’s lightweight MolYOLO detector and the substantially shorter output format enabled by removing verbose bounding-box coordinates.

\subsubsection{Analysis of Different Reaction Layout}

\begin{figure}[h]
    \centering
    \includegraphics[width=0.8\linewidth]{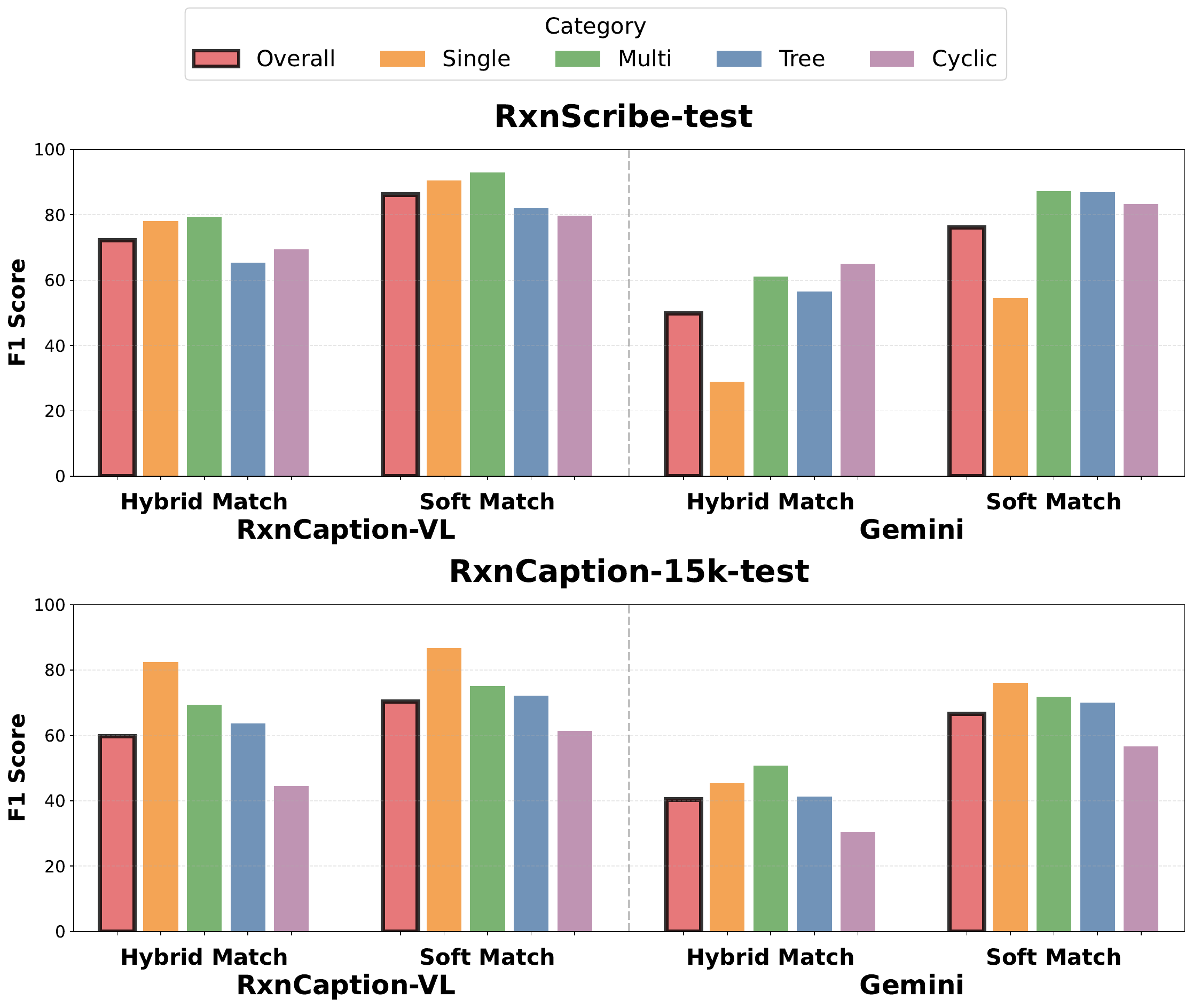}
    \caption{Performance of RxnCaption-VL and Gemini on Different Reaction Layout.} 
    \label{fig:category_comparison}
\end{figure}

Figure~\ref{fig:category_comparison} highlights distinct patterns across reaction layout categories. Across reaction layout categories, we observe clear performance stratification, especially on \texttt{RxnCaption-15k-test}: single-line, multi-line and tree layouts show no substantial performance difference, while cyclic reactions exhibit drops. This progressive pattern highlights that structural complexity remains the primary bottleneck, suggesting substantial room for improvement in modeling complex reaction layouts.

Notably, despite the increased strictness of HybridMatch, our model maintains stable performance after training without experiencing degradation. This stability indicates that our model preserves structural consistency even under stringent matching criteria, further demonstrating its robustness to varying reaction complexities.

\section{Conclusion}

This study presents the RxnCaption framework, enhancing LVLM performance in parsing chemical reaction diagrams by reformulating the task as image captioning through the ``BBox and Index as Visual Prompt'' (BIVP) strategy. Powered by our molecular structure detector MolYOLO, which achieves top accuracy on public benchmarks, BIVP provides a reliable visual prior. We developed the \texttt{RxnCaption-15k} dataset, a manually annotated collection of reaction diagrams from real chemical literature, significantly larger than existing datasets. Using this data and strategy, we trained the RxnCaption-VL model, setting new records in chemical reaction diagram parsing. Our methods, data, and models are expected to advance large-scale, high-precision extraction of reaction information from chemical literature, providing crucial technological support for the development of AI in chemistry.
\section{Limitation}

Despite the rapid advancements in LVLM and agent technologies, extracting structured chemical reaction information from literature remains a complex and highly challenging task. From the perspective of human reading and understanding of chemical literature, this involves a multifaceted process requiring multimodal information processing, specialized knowledge, and deep reasoning and reflection. This paper focuses on the parsing of chemical reaction diagrams, highlighting two limitations:

\begin{enumerate}
    \item \textbf{Visual Parsing Limitations}: The current approach primarily analyzes reaction diagrams from a visual standpoint, lacking the integration of chemical knowledge for collaborative parsing. For instance, authors sometimes place reactants above horizontal arrows connecting reactants and products. This is why the RxnScribe's SoftMatch strategy merges molecular components from conditions into reactants during evaluation. Such scenarios are difficult to accurately parse using only visual methods, which depend on the model learning the spatial relationships between molecular components and symbols like arrows and plus signs. Future research directions include incorporating molecular structure and atomic composition information for more accurate and professional parsing.
    
    \item \textbf{Unresolved Challenges in Reaction Diagram Parsing}: The task of parsing chemical reaction diagrams is not yet fully solved. Although models show generally high metrics on the RxnScribe dataset, the best-performing RxnCaption-VL model achieves  an F1 score of less than 60\% on our proposed \texttt{RxnCaption-15k-test} dataset. Continually improving model accuracy and performance is a key research focus. Additionally, effectively evaluating model uncertainty in challenging or currently unsolvable scenarios, enabling the model to abstain when necessary, is crucial for ensuring the reliability of output results and for facilitating the practical application of this technology.
    
\end{enumerate}

\section{Acknowledgments}
This research was supported by Shanghai Artificial Intelligence Laboratory.

% \section{Acknowledgments}
% \label{section:ack}

\clearpage
\newpage
\bibliographystyle{plainnat}
\setcitestyle{numbers}
\bibliography{paper}

\clearpage
\newpage
\beginappendix

\section{Related Work}
\subsection{Chemical Reaction Mining}

Existing research on extracting chemical reactions from literature primarily employs text mining to identify entities such as reactants and products in patents~\cite{schneider2016big}. Tools like LeadMine perform entity recognition, while fingerprint-based methods enhance role assignment in noisy data~\cite{schneider2016s}. Rule-based and machine learning approaches, including Naive Bayes, are used for section identification and dataset construction~\cite{schneider2016big}. Recent initiatives like the ORD offer structured schemas for reaction data sharing~\cite{kearnes2021open}. Reaction Miner~\cite{zhong2023reaction} and automated methods using Transformer-based ChemBERT improve product identification and role labeling~\cite{guo2021automated}. ReactIE~\cite{zhong2023reactie} employs weak supervision with linguistic cues to enhance extraction. LLMs facilitate zero-shot NER, IUPAC-to-SMILES conversion, and atom mapping, adding 26\% new reactions from patents~\cite{vangala2024suitability}.

Chemical reaction diagram parsing has advanced recently. ReactionDataExtractor 2.0~\cite{wilary2023reactiondataextractor} combines molecular and arrow detectors with text recognition, but its rule-based system struggles with generalization due to cumulative errors. RxnScribe~\cite{qian2023rxnscribe} redefines diagram parsing as a sequence generation task using Pix2Seq~\cite{chen2021pix2seq}, creating a RxnDP dataset from real papers. RxnIm~\cite{RxnIm} introduces LVLMs, using synthetic data to address scarcity and employing a three-stage training strategy: 1) Pre-localization for molecular detection; 2) Full parsing with synthetic data; 3) Fine-tuning with real data. Despite high computational demands, it only marginally improves parsing accuracy over RxnScribe and struggles with out-of-distribution samples (as detailed in \S~4.2 of the Main paper).

\subsection{LVLM and Visual Prompt}

Recently, Large Vision-Language Models (LVLMs)~\cite{hurst2024gpt, bai2023qwenvlversatilevisionlanguagemodel, chen2024internvl} built on Large Language Models (LLMs)~\cite{bai2023qwentechnicalreport, shao2022internnewlearningparadigm, touvron2023llama} have made significant strides. These models excel in visual perception~\cite{Liu_2024}, visual question answering (VQA)~\cite{yu2024mmvetevaluatinglargemultimodal}, and multimodal reasoning~\cite{yue2024mmmuprorobustmultidisciplinemultimodal, gao2025pm4benchparallelmultilingualmultimodal}, with applications in medicine~\cite{li2023llavamedtraininglargelanguageandvision}, autonomous driving~\cite{duan2024cityllavaefficientfinetuningvlms}, remote sensing~\cite{pang2024vhmversatilehonestvision, muhtar2024lhrsbotempoweringremotesensing}, and OCR~\cite{wei2024generalocrtheoryocr20}.

Visual prompts~\cite{wu2024visual} provide input in visual forms (e.g., images, selections, markings) in LVLMs, enhancing fine-grained understanding and task adaptability. Unlike text prompts, visual prompts directly affect the input image, improving visual attention and reducing hallucinations. Set-of-Mark (SoM)~\cite{yang2023set} demonstrated the effectiveness of "discrete markings" by overlaying symbols on images to guide GPT-4V in tasks like directional question-answering without fine-tuning, establishing a "marking as prompt" zero-shot paradigm. Vip-llava~\cite{cai2024vip} allows models to understand arbitrary visual prompts, enabling region-specific question-answering without additional fine-tuning, and introduces the ViP-Bench for evaluation. Image-of-Thought (IoT)~\cite{zhou2024image} suggests models plan a multi-step visual-text reasoning chain, dynamically generating necessary masks or text markings using external tools (e.g., SAM~\cite{kirillov2023segment}, OCR), achieving an automated ``prompt as program" process.
\clearpage
\newpage
\section{Details for Pilot Study}
\subsection{Introduction of RxnCaption-15k}

\begin{table*}[h!]
\centering
\setlength{\tabcolsep}{8pt}
\renewcommand{\arraystretch}{1.2}
\renewcommand{\heavyrulewidth}{1.2pt}
\resizebox{\textwidth}{!}{ % ← 让表格宽度自适应页面
\begin{tabular}{@{}lccc@{}}
\toprule
\textbf{VQA Question} & \textbf{Gemini-2.5-Pro} & \textbf{GPT4o-2024-11-20} & \textbf{QwenVL-Max} \\
\midrule
How many reactions are in the image? & 72.72 & 61.81 & 74.54 \\
How many molecular structures are in the images? & 66.36 & 25.45 & 56.36 \\
Are there any cyclic reactions? & 91.82 & 87.27 & 77.27 \\
Are there any tree-structured reactions? & 72.73 & 71.82 & 86.36 \\
\midrule
Average & 75.91 & 61.59 & 73.63 \\
\bottomrule
\end{tabular}
}
\caption{Zero-shot results of VQA task on RxnScribe-test-slct for General-purpose VLM}
\label{tab:vqa_results}
\end{table*}

\begin{table}[t!]
\centering
\begin{minipage}{0.95\linewidth}
\rule{\linewidth}{0.5pt}\\[0.5ex]
\fontsize{9pt}{11pt}\selectfont
\textbf{Prompt: BROS pattern} \\[0.3ex]
\rule{\linewidth}{0.5pt}

You are given an image containing one or more chemical reaction equations. Each equation has three parts: reactants, conditions, and products. Each part may include multiple objects, and each object is either a structure, text, identifier, or supplement. Please extract all objects and their bounding boxes, and return them in the following strict JSON format: a list, where each element represents a reaction with keys 'reactants', 'conditions', and 'products'. Each key maps to a list of objects, and each object has:

- category: one of "structure", "text", "identifier", or "supplement"

- bbox: a list of four normalized values [x1, y1, x2, y2], representing the bounding box of the object relative to the image size. Each value must be between 0 and 1.

The coordinates must satisfy: x1 \textless x2, y1 \textless y2. (x1, y1) corresponds to the top-left corner, and (x2, y2) to the bottom-right corner of the object in the image.

\begin{verbatim}
[{
  "reactants": [{"category": "structure", 
        "bbox": [0.1, 0.2, 0.3, 0.4]}],
  "conditions": [{"category": "text", 
        "bbox": [0.32, 0.21, 0.4, 0.25]}],
  "products": [{"category": "structure", 
        "bbox": [0.45, 0.2, 0.6, 0.4]}]
}]
\end{verbatim}

Output only the JSON. Do not include any explanation or additional text.

\textless image\textgreater Now output your JSON format result:

\rule{\linewidth}{0.5pt}
\end{minipage}
\caption{Prompt for BROS pattern}
\label{tab:prompt_bros}
\end{table}
\begin{table}[t!]
\centering
\begin{minipage}{0.95\linewidth}
\rule{\linewidth}{0.5pt}\\[0.5ex]
\fontsize{9pt}{11pt}\selectfont
\textbf{Prompt: Reaction Counting} \\[0.3ex]
\rule{\linewidth}{0.5pt}

You are given an image containing one or more chemical reaction equations. Your task is to count how many distinct reaction equations are present in the image. Output a JSON object with a single key 'reaction\_count' and an integer value indicating the number of reactions. For example: \{"reaction\_count": 2\}.  

Do not include any explanation or additional text.

\textless image\textgreater Now output your JSON format result:

\rule{\linewidth}{0.5pt}
\end{minipage}
\caption{Prompt template for counting reaction}
\label{tab:prompt_reaction_counting}
\end{table}
\begin{table}[t!]
\centering
\begin{minipage}{0.95\linewidth}
\rule{\linewidth}{0.5pt}\\[0.5ex]
\fontsize{9pt}{11pt}\selectfont
\textbf{Prompt: Molecular Structure Counting} \\[0.3ex]
\rule{\linewidth}{0.5pt}

You are given an image containing one or more chemical reaction equations. Your task is to count how many distinct molecule structures (i.e., chemical structure diagrams) are present in the image. Output a JSON object with a single key 'structure\_count' and an integer value indicating the number of molecular structures. For example: \{"structure\_count": 4\}.  

Do not include any explanation or additional text.

\textless image\textgreater Now output your JSON format result:

\rule{\linewidth}{0.5pt}
\end{minipage}
\caption{Prompt template for counting molecular structures}
\label{tab:prompt_structure_counting}
\end{table}
\begin{table}[h!]
\centering
\setlength{\tabcolsep}{6pt}
\renewcommand{\arraystretch}{1}
\renewcommand{\heavyrulewidth}{1.2pt} 
\begin{tabular}{@{}c *{3}{c} *{3}{c}@{}}
\toprule
\multirow{2}{*}{\bfseries Model} & 
\multicolumn{3}{c}{\bfseries Hard Match} & 
\multicolumn{3}{c}{\bfseries Soft Match} \\
\cmidrule(lr){2-4} \cmidrule(lr){5-7}
 & \textbf{P} & \textbf{R} & \textbf{F1} & \textbf{P} & \textbf{R} & \textbf{F1} \\
\midrule
\textbf{Gemini-2.5-Pro} & 0.7 & 0.6 & 0.6 & 38.2 & 33.0 & 35.4 \\
\textbf{GPT4o-2024-11-20} & 0.4 & 0.3 & 0.3 & 0.4 & 0.3 & 0.3 \\
\textbf{QwenVL-Max} & 0 & 0 & 0 & 5.0 & 4.0 & 4.4 \\
\bottomrule
\end{tabular}
\caption{Zero-shot results of RxnDP task on RxnScribe-test-slct for General-purpose VLM in BROS strategy}
\label{tab:bros_results}
\end{table}

\begin{table}[h!]
\centering
\setlength{\tabcolsep}{5pt}
\renewcommand{\arraystretch}{1.1}
\renewcommand{\heavyrulewidth}{1.2pt} 
\begin{tabular}{@{}c *{3}{c} *{3}{c}@{}}
\toprule
\multirow{2}{*}{\bfseries Model} & 
\multicolumn{3}{c}{\bfseries Hard Match} & 
\multicolumn{3}{c}{\bfseries Soft Match} \\
\cmidrule(lr){2-4} \cmidrule(lr){5-7}
 & \textbf{P} & \textbf{R} & \textbf{F1} & \textbf{P} & \textbf{R} & \textbf{F1} \\
\midrule
\textbf{Gemini-2.5-Pro} & 17.2 & 18.8 & 17.9 & 77.8 & 84.5 & 81.0 \\
\textbf{GPT4o-2024-11-20} & 9.9 & 9.7 & 9.8 & 59.7 & 55.7 & 57.6 \\
\textbf{QwenVL-Max} & 14.5 & 13.4 & 13.9 & 69.6 & 63.6 & 66.5 \\
\bottomrule
\end{tabular}
\caption{Zero-shot results of RxnDP task on RxnScribe-test-slct for General-purpose VLM in BIVP strategy}
\label{tab:bivp_results}
\end{table}
\begin{table}[t!]
\centering
\begin{minipage}{0.95\linewidth}
\rule{\linewidth}{0.5pt}\\[0.5ex]
\fontsize{9pt}{11pt}\selectfont
\textbf{Prompt: Cyclic Reaction Detection} \\[0.3ex]
\rule{\linewidth}{0.5pt}

You are given an image of a chemical reaction. Your task is to determine whether the image contains a cyclic (graph-style) chemical reaction diagram. Output a JSON object with a single key 'cyclic' and a boolean value indicating whether the image contains a cyclic (graph-style) chemical reaction diagram. If the image contains a cyclic reaction, output: \{"cyclic": true\}. Otherwise, output: \{"cyclic": false\}.  

Do not include any explanation or additional text.

\textless image\textgreater Now output your JSON format result:

\rule{\linewidth}{0.5pt}
\end{minipage}
\caption{Prompt template for detecting cyclic chemical reactions}
\label{tab:prompt_cyclic_detection}
\end{table}
\begin{table}[t!]
\centering
\begin{minipage}{0.95\linewidth}
\rule{\linewidth}{0.5pt}\\[0.5ex]
\fontsize{9pt}{11pt}\selectfont
\textbf{Prompt: Tree Reaction Detection} \\[0.3ex]
\rule{\linewidth}{0.5pt}

You are given an image of a chemical reaction. Your task is to determine whether the image contains a tree (tree-style) chemical reaction diagram. Output a JSON object with a single key 'tree' and a boolean value indicating whether the image contains a tree (tree-style) chemical reaction diagram. If the image contains a tree reaction, output: \{"tree": true\}. Otherwise, output: \{"tree": false\}.  
Do not include any explanation or additional text.  \\
\textless image\textgreater~Now output your JSON format result: \\
\rule{\linewidth}{0.5pt}
\end{minipage}
\caption{Prompt template for detecting tree-structured chemical reaction}
\label{tab:prompt_tree_detection}
\end{table}

\begin{table}[t!]
\centering
\begin{minipage}{0.95\linewidth}
\rule{\linewidth}{0.5pt}\\[0.5ex]
\fontsize{9pt}{11pt}\selectfont
\textbf{Prompt: BIVP pattern} \\[0.3ex]
\rule{\linewidth}{0.5pt}

You are an expert in chemical image structure analysis. You will be given an image in which all "molecular structures" have been boxed and numbered. Your task is to identify and reconstruct the chemical reaction(s) in the image based on these boxed structures. Please follow the rules below:

1. Each reaction must contain three fields: reactants, conditions, products (each is a List[Dict])  \\
2. Each Dict must include:  \\
- type: one of "mol", "txt", or "idt"  \\
    - if type is "mol", second field is \texttt{index} (box ID);  \\
    - if type is "txt" or "idt", second field is \texttt{content} (raw text)  \\
3. Boxed molecular structures must be type "mol". Other elements must use "txt" or "idt"  \\
4. Ignore decorations, arrows, and illustrations; only extract real elements  \\
5. Clean up identifiers and retain only the core ID  \\
6. If no reaction exists in the image, return an empty list []\\  
7. Use arrow direction: tail = reactants, head = products\\

Example format:

\begin{verbatim}
[
  {
    "reactants": [
    {"type": "mol", "index": 1}, 
    {"type": "txt", "content": "NaCl"}
    ],
    "conditions": [
    {"type": "mol", "index": 3}, 
    {"type": "txt", "content": "H2O, 25°C"}
    ],
    "products": [
    {"type": "mol", "index": 2}, 
    {"type": "idt", "content": "1a"}
    ]
  }
]
\end{verbatim}

Do not include any explanations, comments, or non-structured data.\\
\textless image\textgreater~Please extract the list of reactions from the image. Return only the JSON reaction list without adding any other content.
\rule{\linewidth}{0.5pt}
\end{minipage}
\caption{Prompt for BIVP pattern}
\label{tab:prompt_bivp}
\end{table}

We define the Reaction Diagram Parsing (RxnDP) task as \(
R = \mathcal{F}(I)
\),
where \( I \) represents the input chemical reaction diagram, \( \mathcal{F} \) is the RxnDP model, and \(R = \{ R_i \}\) is the set of all chemical reactions in the diagram.

Each reaction \(R_i\) consists of three roles:  
\[
R_i=\bigl(\underset{\text{reactants}}{\mathcal C_{\text{react}}},\;
\underset{\text{conditions}}{\mathcal C_{\text{cond}}},\;
\underset{\text{products}}{\mathcal C_{\text{prod}}}\bigr)
\]

Each role \(\mathcal{C}_{\text{role}} = \{ c_j \}\) (\(\text{role} \in \{\text{react}, \text{cond}, \text{prod}\}\)) is composed of several components. The components can be in two modalities: molecular structure diagrams and text. A molecular structure diagram component (referred to as molecular component) is represented by the bounding box coordinates of the molecule in the diagram, while a text component is represented directly by its textual content.

\begin{figure*}
    \centering
    \includegraphics[width=1.0\linewidth]{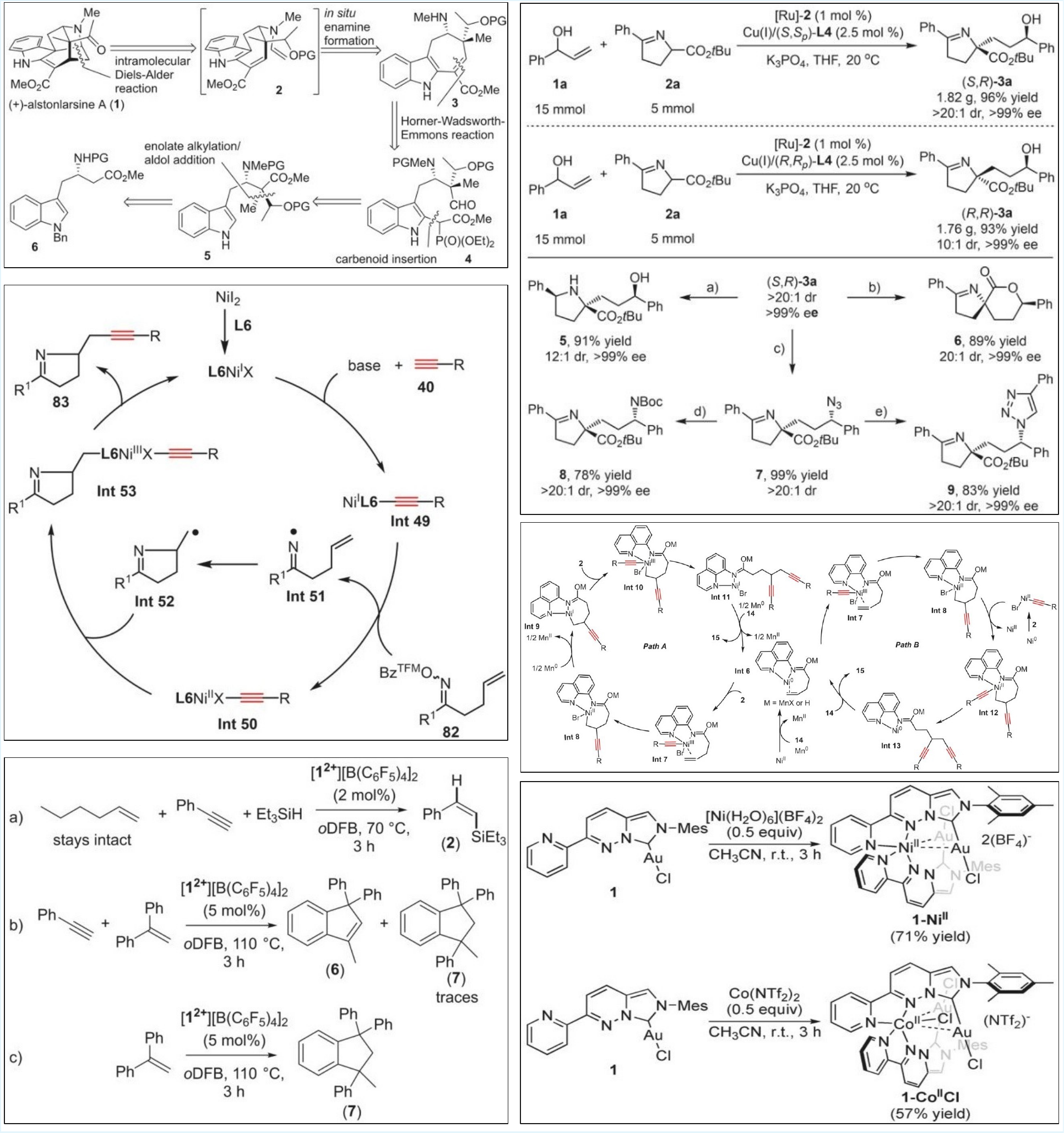}
    \caption{Visualization results of RxnCaption-15k's training data.}
    \label{fig:RxnDP}
\end{figure*}

Existing RxnDP datasets have notable limitations: the \texttt{RxnScribe} dataset, derived from real papers, includes only 1378 samples, while the \texttt{RxnIm} dataset, though large, is artificially synthesized, limiting diversity and model generalization (see \S~4.2 of the Main paper). To overcome these issues, we developed the \texttt{RxnCaption-15k} dataset—a large, high-quality, and diverse RxnDP dataset. And the training data of RxnCaption-15k is shown in Figure~\ref{fig:RxnDP}.

\subsection{Prompts for LVLM inference}

The Prompt template for BROS pattern is shown in Table ~\ref{tab:bros_results}.
The prompt template for VQA are shown in Table ~\ref{tab:prompt_reaction_counting}, ~\ref{tab:prompt_structure_counting}, ~\ref{tab:prompt_cyclic_detection}, ~\ref{tab:prompt_tree_detection}.
The prompt template for BIVP is shown in Table ~\ref{tab:bivp_results}.

\subsection{Metric for Pilot Study}
\textbf{HardMatch}: A strict reaction-level matching strategy that requires all components of a reaction — including molecular structures and textual elements in reactants, products, and conditions — to be correctly matched. Each predicted component must have an IoU \textgreater 0.5 with its ground truth counterpart. A reaction is counted as a true positive only if all associated components meet this criterion.

\textbf{SoftMatch}: A relaxed reaction-level matching strategy that ignores all text components and merges conditions into reactants during evaluation. A reaction is counted as a true positive only if all molecular structures in the reactants and products achieve IoU \textgreater 0.5 with the ground truth. This strategy focuses solely on structural alignment.

\subsection{Detailed Results for Pilot Study}

The results of RxnDP task on RxnScribe-test-slct for General-purpose VLM in BROS strategy, VQA task, and RxnDP task in BIVP strategy are shown in Tables~\ref{tab:bros_results},~\ref{tab:vqa_results}, and~\ref{tab:bivp_results}, respectively.
\clearpage
\newpage
\section{Details of method}

\subsection{OCR details for Chemical Diagram Parsing}

For the RxnCaption-15k dataset, each image contains annotated bounding boxes (bbox) that identify specific regions of interest within chemical diagrams. We extract the bounding box coordinates and crop the corresponding image regions, then apply a specialized OCR prompt(see Table~\ref{tab:prompt_ocr}) to each cropped region using Gemini-2.5-Pro.

\begin{table}[h]
\centering
\begin{minipage}{0.95\linewidth}
\rule{\linewidth}{0.5pt}\\[0.5ex]
\fontsize{9pt}{11pt}\selectfont
\textbf{Prompt: Gemini OCR} \\[0.3ex]
\rule{\linewidth}{0.5pt}

\begin{quote}
OCR\_PROMPT = "You are a master OCR assistant for chemistry, tasked with analyzing an image snippet from a chemical diagram. Your goal is to provide the most accurate text representation.

\textbf{Follow these steps:}

\textbf{Step 1:} Analyze the image content type.
\begin{itemize}
\item Is it a graphical chemical structure? (e.g., contains bond lines, benzene rings, complex drawings).
\item Is it purely text? (e.g., can be typed on a keyboard).
\end{itemize}

\textbf{Step 2:} Provide output based on the content type.
\begin{itemize}
\item If it is a graphical structure: You MUST return the special token [GRAPHICAL\_STRUCTURE]. Do not attempt to OCR it.
\item If it is purely text: Transcribe the text with maximum precision.
   1. Pay attention to details: Preserve case-sensitivity (e.g., Pd vs pd), all symbols (degree, /, -, +, (), and numbers.
   2. Handle multi-line text: If text is on multiple lines, combine them into a single line separated by a space.
\end{itemize}

\textbf{Examples:}
\begin{itemize}
\item Image shows a drawing of a benzene ring \textrightarrow Output: [GRAPHICAL\_STRUCTURE]
\item Image shows a drawing of a molecule labeled "1a" \textrightarrow Output: [GRAPHICAL\_STRUCTURE]
\item Image shows the text (CH3)2N \textrightarrow Output: (CH3)2N
\item Image shows the text Pd/C, H2 \textrightarrow Output: Pd/C, H2
\item Image shows the text electrocatalyst \textrightarrow Output: electrocatalyst
\item Image shows text on two lines: NaH and THF \textrightarrow Output: NaH THF
\end{itemize}

\textbf{Final Output Format:}
Return ONLY the transcribed text OR the [GRAPHICAL\_STRUCTURE] token. Do not add explanations, markdown, or quotes."
\end{quote}

Output only the JSON. Do not include any explanation or additional text.  
\textless image\textgreater Now output your JSON format result:

\rule{\linewidth}{0.5pt}
\end{minipage}
\caption{Prompt for Gemini OCR}
\label{tab:prompt_ocr}
\end{table}

\subsection{Details of four layouts}

\begin{figure*}
    \centering
    \includegraphics[width=1.0\linewidth]{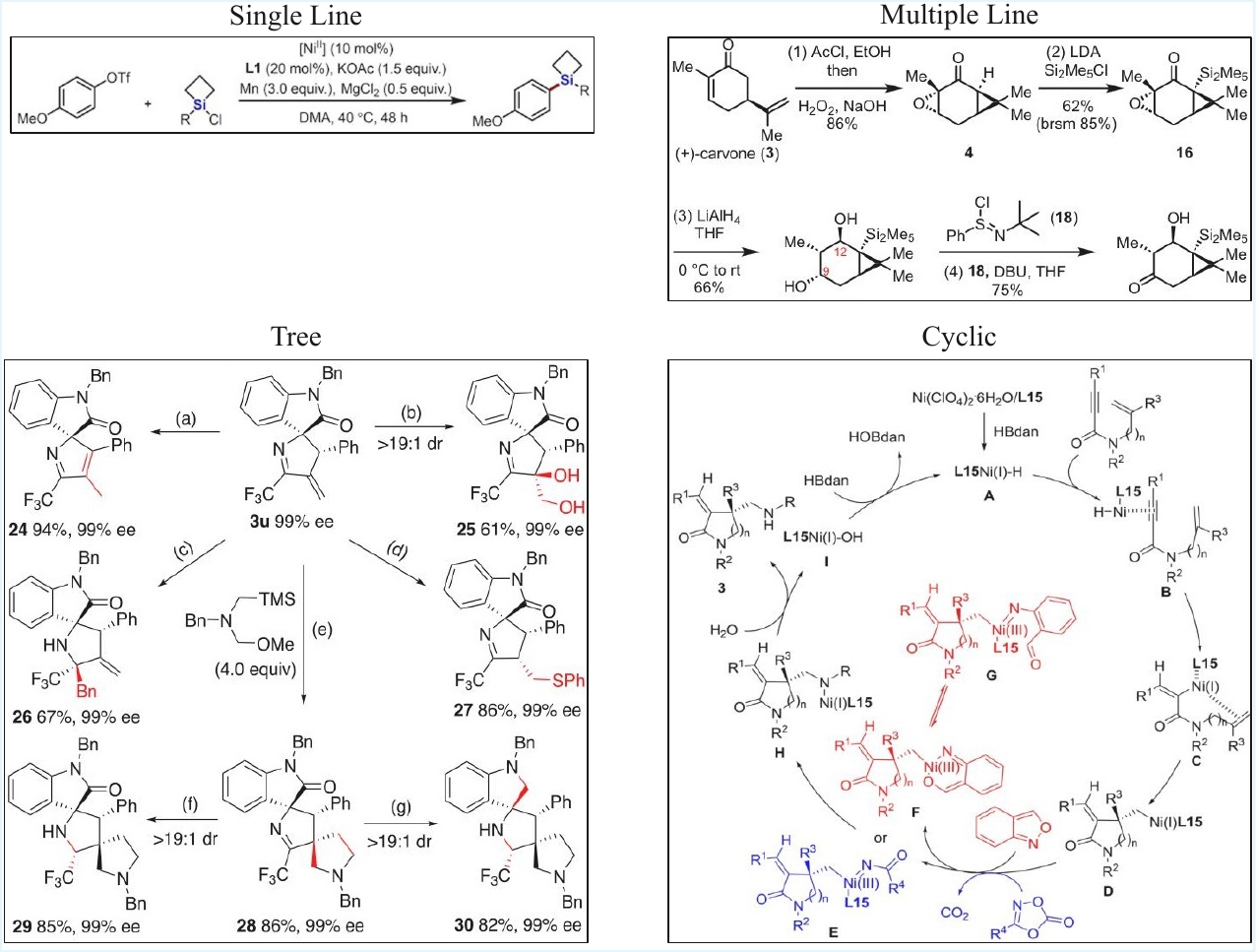}
    \caption{Four layouts of Rxn Diagram.}
    \label{fig:four_layouts}
\end{figure*}

The visualization of chemical reaction processes typically follows specific logical layouts to accurately convey their intrinsic chemical relationships. There are four primary types of layouts, as shown in Figure \ref{fig:four_layouts}: 1) the Single-line Layout, which is characterized by the linear display of a series of continuous chemical transformations along a single path, sequentially arranging starting materials, intermediates, and products to form an unbroken reaction chain; 2) the Multiple-line Layout, which structurally breaks down a long reaction route into several consecutive linear segments arranged in sections; 3) the Tree Layout, which features multiple independent reaction pathways diverging from a common starting material or intermediate to different products, forming a one-to-many radial structure; and 4) the Cyclic Layout, which uses a closed-loop diagram to describe the process of a species undergoing a series of transformations and ultimately regenerating to its initial state.

\clearpage
\newpage

\section{Details of the experiments}

\subsection{Results Categorized into Four Layout Types}

% We conducted a detailed analysis on four layout categories using four trained models from the main experiment table and one top-performing zero-shot VLM, Gemini-2.5-Pro. The results are shown in Figure~\ref{fig:layout_rxnscribe} and Figure~\ref{fig:layout_pdf}. From the figures, we can observe that, when categorized by layout type, the difficulty of the RxnDP task follows the order: \textbf{Single-line $<$ Multi-line $<$ Tree $<$ Cyclic}. Performance across all categories is generally slightly lower on \texttt{RxnCaption-15k-test}, which is due to our image crop being performed at the figure/table level, resulting in greater diversity and realism. In terms of individual model performance, our \textbf{RxnCaption-VL} demonstrates more balanced results across all categories, especially on \texttt{RxnScribe-test}, whereas Gemini-2.5-Pro exhibits a more pronounced imbalance in performance across different categories.

\begin{figure*}[t]
    \centering
    \includegraphics[width=\linewidth]{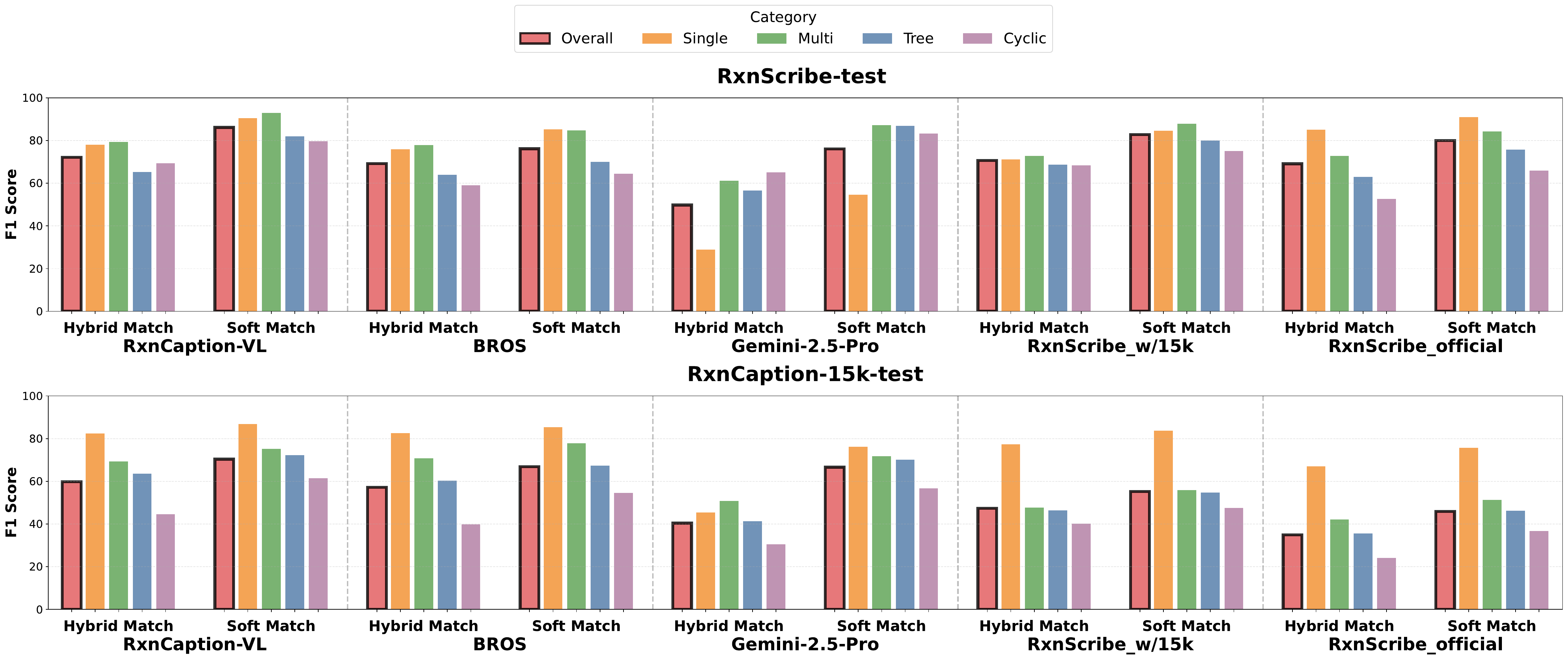}
    \caption{Results categorized into four layout types on \texttt{RxnScribe-test} and \texttt{RxnCaption-15k-test}.}
    \label{fig:category_comparison_appendix}
\end{figure*}

We conducted a detailed analysis on four layout categories using four trained models from the main experiment table and one top-performing zero-shot VLM, Gemini-2.5-Pro. The results are shown in Figure~\ref{fig:category_comparison_appendix}. From the figures, we can observe that, when categorized by layout type, the difficulty of the RxnDP task follows the order: \textbf{Single-line $<$ Multi-line $<$ Tree $<$ Cyclic}. Performance across all categories is generally slightly lower on \texttt{RxnCaption-15k-test}, which is due to our image crop being performed at the figure/table level, resulting in greater diversity and realism. In terms of individual model performance, our \textbf{RxnCaption-VL} demonstrates more balanced results across all categories, especially on \texttt{RxnScribe-test}, whereas Gemini-2.5-Pro exhibits a more pronounced imbalance in performance across different categories.

Besides, by comparing the pix2seq-based models (RxnScribe-official and RxnScribe\_w/15k) with the VLM-trained models, we observe that traditional approaches maintain relatively stable performance on categories such as single-line and multi-line, but experience a significant drop when handling more complex images like tree and cyclic structures. This suggests that the VLM’s strong visual-language understanding capability can greatly benefit reaction diagram parsing tasks involving intricate graph structures. 
Moreover, by comparing the performance of RxnScribe-official and RxnScribe\_w/15k across the two test sets, we observe that our proposed \texttt{RxnCaption-15k} not only yields substantial improvements on its corresponding test set, but also leads to noticeable gains on the \texttt{RxnScribe-test}. This demonstrates the strong generalizability of our constructed training dataset.
% \begin{figure}[H]
%     \centering
%     \includegraphics[width=1.0\linewidth]{Appendix_Figures/layout_rxnscribe.pdf}
%     \caption{Results categorized into four layout types on \texttt{RxnScribe-test}.}
%     \label{fig:layout_rxnscribe}
% \end{figure}
% \begin{figure}[H]
%     \centering
%     \includegraphics[width=1.0\linewidth]{Appendix_Figures/layout_pdf.pdf}
%     \caption{Results categorized into four layout types on \texttt{RxnCaption-15k-test}.}
%     \label{fig:layout_pdf}
% \end{figure}

\subsection{Detailed Results of Ablation Studies}

For ablation experiments \textbf{Error Analysis of BIVP Strategy} and \textbf{Influence of Molecular Detector} in the main text, we present the detailed results in Table~\ref{tab:BIVP_error} and Table~\ref{tab:Molecular Detector}, which include the Precision, Recall, and F1 scores for each dataset and each evaluation metric.

\clearpage 

\begin{table*}[h!]
\centering
\setlength{\tabcolsep}{6pt}
\renewcommand{\arraystretch}{1}
\renewcommand{\heavyrulewidth}{1.2pt} 
\resizebox{\textwidth}{!}{ % ← 让表格宽度自适应页面
\begin{tabular}{@{}c c c *{6}{c} *{6}{c}@{}}
\toprule
\multirow{4}{*}{\bfseries Detector Setup} & 
\multirow{4}{*}{\bfseries Rxn Extractor} &
\multicolumn{6}{c}{\bfseries \texttt{RxnScribe-test}} & 
\multicolumn{6}{c}{\bfseries \texttt{RxnCaption-15k-test}} \\
\cmidrule(lr){3-8} \cmidrule(lr){9-14}
 &  & \multicolumn{3}{c}{\bfseries Soft Match} & \multicolumn{3}{c}{\bfseries Hybrid Match} & \multicolumn{3}{c}{\bfseries Soft Match} & \multicolumn{3}{c}{\bfseries Hybrid Match} \\
\cmidrule(lr){3-5} \cmidrule(lr){6-8} \cmidrule(lr){9-11} \cmidrule(lr){12-14}
 &  & \textbf{P} & \textbf{R} & \textbf{F1} & \textbf{P} & \textbf{R} & \textbf{F1} & \textbf{P} & \textbf{R} & \textbf{F1} & \textbf{P} & \textbf{R} & \textbf{F1} \\

\midrule
\multirow{2}{*}{\textbf{MolYOLO}} & Gemini-2.5-pro & 67.9 & 86.5 & 76.1 & 44.7 & 56.1 & 49.8 & 64.2 & 69.2 & 66.2 & 38.9 & 42.1 & 40.4 \\
 & RxnCaption-VL & 85.3 & 87.1 & 86.2 & 71.6 & 72.7 & 72.2 & 71.3 & 69.4 & 70.4 & 60.3 & 59.3 & 59.8 \\
 \multirow{2}{*}{\makecell{\textbf{GT bbox + } \\ \textbf{MolYOLO Recall}}} & Gemini-2.5-pro & 74.5 & 91.7 & 82.2 & 45.9 & 58.7 & 51.5 & 70.6 & 76.6 & 73.5 & 44.7 & 48.3 & 46.4 \\
 & RxnCaption-VL & 87.2 & 88.1 & 87.6 & 73.3 & 73.5 & 73.4 & 75.9 & 76.7 & 76.3 & 63.8 & 63.9 & 63.8 \\
\textbf{MolYOLO} & Ground Truth & 100.0 & 99.4 & 99.7 & 100.0 & 99.4 & 99.7 & 100.0 & 91.0 & 95.3 & 100.0 & 91.0 & 95.3 \\

\bottomrule
\end{tabular}
}
\caption{Detailed results (Table 6 in full paper) of Error Attribution Analysis of BIVP Strategy in ablation studies.}
\label{tab:BIVP_error}
\end{table*}

\begin{table*}[h!]
\centering
\setlength{\tabcolsep}{6pt}
\renewcommand{\arraystretch}{1}
\renewcommand{\heavyrulewidth}{1.2pt} 
\resizebox{\textwidth}{!}{ % ← 让表格宽度自适应页面
\begin{tabular}{@{}c c c *{6}{c} *{6}{c}@{}}
\toprule
\multirow{4}{*}{\bfseries LVLM} & 
\multirow{4}{*}{\makecell{\textbf{Molecular}\\\textbf{Detector}}} &
\multicolumn{6}{c}{\bfseries \texttt{RxnScribe-test}} & 
\multicolumn{6}{c}{\bfseries \texttt{RxnCaption-15k-test}} \\
\cmidrule(lr){3-8} \cmidrule(lr){9-14}
 &  & \multicolumn{3}{c}{\bfseries Soft Match} & \multicolumn{3}{c}{\bfseries Hybrid Match} & \multicolumn{3}{c}{\bfseries Soft Match} & \multicolumn{3}{c}{\bfseries Hybrid Match} \\
\cmidrule(lr){3-5} \cmidrule(lr){6-8} \cmidrule(lr){9-11} \cmidrule(lr){12-14}
 &  & \textbf{P} & \textbf{R} & \textbf{F1} & \textbf{P} & \textbf{R} & \textbf{F1} & \textbf{P} & \textbf{R} & \textbf{F1} & \textbf{P} & \textbf{R} & \textbf{F1} \\

\midrule
\multirow{3}{*}{\textbf{Gemini-2.5-pro}} & \textbf{YoDe} & 49.4 & 60.7 & 54.5 &  30.1 & 41.0 & 34.7 & 58.9 & 59.9 & 59.4 & 35.0 & 37.1 & 36.0 \\
& \textbf{MolDetect} & 64.3 & 82.6 & 72.3 & 42.5 & 54.9 & 47.9 & 59.2 & 62.7 & 60.9 & 36.7 & 38.9 & 37.8 \\
 & \textbf{MolYOLO} & 67.9 & 86.5 & 76.1 & 44.7 & 56.1 & 49.8 & 64.2 & 69.2 & 66.6 & 38.9 & 42.1 & 40.4 \\
 \multirow{3}{*}{\textbf{RxnCaption-VL}} & \textbf{YoDe} & 63.4 & 59.7 & 61.5 & 54.3 & 52.4 & 53.3 & 62.2 & 58.3 & 60.2 & 49.9 & 48.5 & 49.2 \\ 
 & \textbf{MolDetect} & 84.1 & 84.7 & 84.4 & 70.9 & 70.7 & 70.8 & 65.3 & 62.8 & 64.0 & 54.0 & 52.5 & 53.2 \\
 & \textbf{MolYOLO} & 85.3 & 87.1 & 86.2 & 71.6 & 72.7 & 72.2 & 71.3 & 69.4 & 70.4 & 60.3 & 59.3 & 59.8 \\

\bottomrule
\end{tabular}
}
\caption{Detailed results (Table 5 in full paper) of Influence of Molecular Detector in ablation studies.}
\label{tab:Molecular Detector}
\end{table*}

% \subsection{Detail of Step 4 in RxnCaption-15k Construction}

% \subsubsection{Prompt of Gemini-2.5-Pro OCR}

% \subsection{Influence of Molecule Detection Score Threshold}

\end{document}